\documentclass{article}

\AtBeginDocument{%
  \providecommand\BibTeX{{%
    \normalfont B\kern-0.5em{\scshape i\kern-0.25em b}\kern-0.8em\TeX}}}

\usepackage[nonatbib,preprint]{neurips_2024}
\usepackage[numbers, sort&compress]{natbib}
\usepackage{comment}

\usepackage{amsmath,amsfonts,bm,bbm}
\usepackage{mathrsfs,amssymb,mathtools}
\usepackage{amsthm}

\def\eqref#1{equation~\ref{#1}}

\def\1{\bm{1}}

\DeclareMathAlphabet{\mathsfit}{\encodingdefault}{\sfdefault}{m}{sl}
\SetMathAlphabet{\mathsfit}{bold}{\encodingdefault}{\sfdefault}{bx}{n}

\newcommand{\R}{\mathbb{R}}

\DeclareMathOperator*{\argmin}{arg\,min}
\DeclareMathOperator*{\minimize}{Minimize}

\newcommand{\cI}{\mathcal{I}}

 \newcommand{\cY}{\mathcal{Y}}
 \newcommand{\cX}{\mathcal{X}}

 \newcommand{\RR}{\mathbb{R}}

\newcommand{\btheta}{{\bm{\theta}}}
\newtheorem{lemma}{Lemma}
\newtheorem{theorem}{Theorem}

\usepackage[utf8]{inputenc} 
\usepackage[T1]{fontenc}    
\usepackage{hyperref}       
\usepackage{url}            
\usepackage{booktabs}       
\usepackage{amsfonts}       
\usepackage{nicefrac}       
\usepackage{microtype}      
\usepackage{xcolor}         

\usepackage{color, colortbl}
\usepackage{paralist}
\usepackage{enumitem}
\usepackage{multirow}
\usepackage{caption}
\usepackage{subcaption}
\usepackage{algorithm}
\usepackage{algorithmic}
\usepackage{makecell}
\usepackage{tikz}
\usepackage{pgfplots}
\usepackage{adjustbox}
\usepackage{wrapfig}
\usepackage{soul}
\usetikzlibrary{positioning}

\newcommand{\pbox}[3]{%
  \begingroup
  \setlength{\fboxsep}{#1} 
  \colorbox{#2}{#3}
  \endgroup
}

\definecolor{darkgreen}{rgb}{0.0, 0.5, 0.0}
\definecolor{darkblue}{rgb}{0.0, 0.0, 0.5}
\definecolor{darkred}{rgb}{0.5, 0.0, 0.0}

\newcommand{\hlgreen}[1]{\!\pbox{1pt}{green!20}{#1}\!}
\newcommand{\hlblue}[1]{\!\pbox{1pt}{cyan!20}{#1}\!}
\newcommand{\hlred}[1]{\!\pbox{1pt}{red!20}{#1}\!}
\newcommand{\mhlblue}[1]{\pbox{3pt}{cyan!20}{$\displaystyle #1$}}
\newcommand{\mhlgreen}[1]{\pbox{3pt}{green!20}{$\displaystyle #1$}}

\newcommand{\bX}{\mathbf{X}}
\newcommand{\bY}{\mathbf{Y}}

\makeatletter
\newcommand{\oset}[3][0ex]{\mathrel{\mathop{#3}\limits^{\vbox to#1{\kern-2\ex@\hbox{$\scriptstyle#2$}\vss}}}}
\makeatother
\newcommand{\optimal}[1]{\oset{\scalebox{.5}{$\star$}}{#1}}

\usepackage{letltxmacro}
\LetLtxMacro\orgvdots\vdots
\LetLtxMacro\orgddots\ddots
\usepackage{amssymb}
\usepackage{pifont}

\definecolor{featsel}{RGB}{0,204,255}
\definecolor{randsub}{RGB}{0,122,255}
\definecolor{perrandsub}{RGB}{7,0,196}
\definecolor{taylor}{RGB}{255,107,107}
\definecolor{metamodel}{RGB}{211,36,48}
\definecolor{genetic}{RGB}{107,0,0}

\definecolor{taylordef}{RGB}{0,204,255}
\definecolor{metamodeldef}{RGB}{211,36,48}
\definecolor{geneticdef}{RGB}{0,217,25}

\title{The Data Minimization Principle in Machine Learning}

\author{%
  Prakhar Ganesh \\
  McGill University \& Mila\\
  \texttt{prakhar.ganesh@mila.quebec} \\
  \And
  Cuong Tran \\
  University of Virginia \& Dyania Health\\
  \texttt{kxb7sd@virginia.edu} \\
  \And
  Reza Shokri \\
  National University of Singapore\\
  \texttt{reza@comp.nus.edu.sg} \\
  \And
  Ferdinando Fioretto \\
  University of Virginia\\
  \texttt{fioretto@virginia.edu} \\
}

\begin{document}

\maketitle

\begin{abstract}
The principle of \emph{data minimization} aims to reduce the amount of data collected, processed or retained to minimize the potential for misuse, unauthorized access, or data breaches. Rooted in privacy-by-design principles, data minimization has been endorsed by various global data protection regulations. However, its practical implementation remains a challenge due to the lack of a rigorous formulation. This paper addresses this gap and introduces an optimization framework for data minimization based on its legal definitions. It then adapts several optimization algorithms to perform data minimization and conducts a comprehensive evaluation in terms of their compliance with minimization objectives as well as their impact on user privacy. Our analysis underscores the mismatch between the privacy expectations of data minimization and the actual privacy benefits, emphasizing the need for approaches that account for multiple facets of real-world privacy risks.
\end{abstract}

\maketitle\allowdisplaybreaks\sloppy

\section{Introduction}
\label{sec:introduction}
The proliferation of data-driven systems and machine learning (ML) applications escalates a number of privacy risks, including those related to unauthorized access to sensitive information~\citep{nasr2019comprehensive,thomas2022framework}.  In response, international data protection frameworks like the European General Data Protection Regulation (GDPR), the California Privacy Rights Act (CPRA), and the Brazilian General Data Protection Law (LGPD) have adopted \emph{data minimization} as a key principle to mitigate these risks~\citep{biegaRevivingPurposeLimitation2021}.

At its core, the data minimization principle requires organizations to \emph{collect, process, and retain only personal data that is adequate, relevant, and limited to what is necessary for specified objectives} (see Table \ref{tab:legal} for further details). It's grounded in the expectation that not all collected data is essential for the objective and, instead, contributes to a heightened risk of information leakage~\citep{goldsteen2021data,paul2021deep,sorscher2022beyond,shanmugam2022learning}. 
However, despite its legal significance and endorsement by global data protection regulations, the data minimization principle lacks a mathematical formalization suitable for real-world ML applications. In particular, as reviewed in \S \ref{sec:related_work}, the current discourse on data minimization practices often overlooks two crucial aspects: {\bf (1)} the individualized nature of minimization (e.g., information that is unimportant for an individual may be critical for another) and {\bf (2)} its intrinsic link to data privacy. 

\textbf{Contributions.} To overcome these limitations, this paper introduces a formal framework for data minimization in ML while being faithful to its legal notion (\S\S \ref{sec:data_minimization}, \ref{sec:formalization}); adapts and evaluates various optimization algorithms to solve the problem of data minimization (\S \ref{sec:min_utility}); and analyzes their compatibility with real-world privacy (\S \ref{sec:min_privacy}). 
In particular, we seek to answer a critical question: \emph{``Do current data minimization requirements in various regulations genuinely meet privacy expectations in legal frameworks?''} Our evaluations reveal that the answer is, unfortunately, no. While being an implicit intention, the requirements of data minimization are not necessarily aligned with risk of reconstruction and re-identification and thus may not provide the expected privacy protection. 
In summary, this paper makes the following contributions:
\begin{enumerate}[leftmargin=*, parsep=0pt, itemsep=0pt, topsep=0pt]
\item It examines various global data protection regulations and provides the first formalization of data minimization for ML tasks that faithfully incorporates the individualized nature of minimization.
\item It compares several algorithms to solve the data minimization problem and conduct an extensive evaluation focusing on key characteristics of the minimized datasets, including emergent individualization and multiplicity.
\item Through extensive evaluations on re-identification and reconstruction attacks on the minimized dataset, it assesses the shortcomings of current regulatory requirements of data minimization in meeting the implicit privacy expectations.
\item Finally, it proposes simple yet effective modifications to the data minimization algorithms and demonstrates that attention to privacy during minimization can provide better trade-offs between user privacy and downstream utility.
\end{enumerate}
This work aims to lay down a path for future research for developing privacy-preserving ML systems in compliance with the legal requirements of data minimization.

\section{The data minimization principle}
\label{sec:data_minimization}

\begin{figure}
    \centering
    \includegraphics[width=0.9\linewidth]{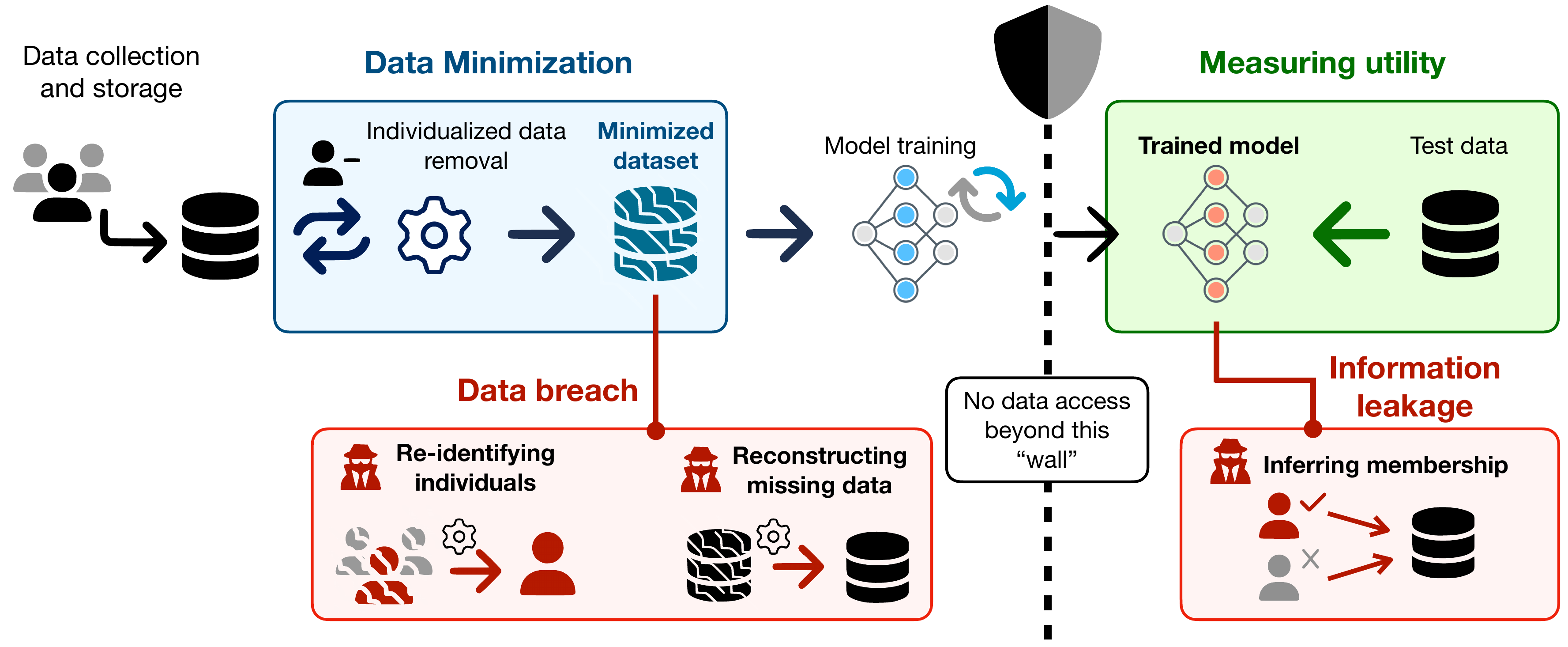}
    \caption{Overview of our framework to study data minimization in a real-world ML pipeline. Behind the data access ``wall'', we highlight the \hlblue{formalization of data minimization} and quantify the risks of a \hlred{data breach}. Outside the ``wall'' with no direct data access, we establish the data minimization objectives using \hlgreen{utility measurement} and study \hlred{information leakage} from the trained model. 
    }
    \label{fig:framework}
    \vspace{-1.5em}
\end{figure}

To faithfully develop the proposed framework (illustrated in Figure~\ref{fig:framework}), we inspect the data minimization principle from six global data protection regulations. 
Building on existing work interpreting the legal language~\citep{biegaRevivingPurposeLimitation2021,quinn2021difficulty,staab2024principle}, we focus instead on operationalizing data minimization, grounded in the common foundations of data privacy and security principles shared by these regulations.
Relevant extracts are summarized in Table \ref{tab:legal}, with complete details reported in \S \ref{app:legal}. The table highlights two core pillars of data minimization: {\bf (i)} {\em purpose limitation}, and {\bf (ii)} {\em data relevance}. 
\begin{itemize}[leftmargin=*, parsep=0pt, itemsep=0pt, topsep=0pt]
    
    \item \hlgreen{\textbf{Purpose limitation}}: 
    Regulations mandate that data be collected for a \textit{legitimate, specific and explicit purpose} (LGPD, Brazil) and prohibit using the collected data for any other incompatible purpose from the one disclosed (CPRA, USA). Thus, data collectors must define a clear, legal objective before data collection and use the data solely for that objective. In machine learning, this \emph{purpose} translates to collecting data solely for training models to achieve optimal performance, where the performance metric needs to be clearly defined~\citep{biegaRevivingPurposeLimitation2021}.

    \item \hlblue{\textbf{Data relevance}}: 
    Regulations like the GDPR require that all collected data {\em be adequate, relevant, and limited to what is necessary for the purposes it was collected for}. In other words, data minimization aims to remove data that does not serve the purpose defined. In ML contexts, this principle translates to retaining only data that contributes to the performance of the model. 
    We interpret this requirement as an optimization objective, minimizing data size while maintaining model performance, under a binary interpretation that the data is either collected or not.
\end{itemize}

\begin{table}[tb!]
    \centering
    \footnotesize
    \begin{tabular}{p{0.95\linewidth}}
        \toprule
        \rowcolor{gray!40}
        \textbf{General Data Protection Regulation (GDPR), Europe} \hfill \href{https://gdpr-info.eu/}{gdpr-info.eu/} \\
        \midrule
        {\sl Article 5(1)(b)}: 
            \textit{Personal data shall be collected for  \hlgreen{specified, explicit and legitimate purposes and not furt-} \hlgreen{her processed in a manner that is incompatible with those purposes}\!;}\\[0.3em]
        {\sl Article 5(1)(c)}: 
            \textit{Personal data shall be \hlblue{adequate, relevant and limited to what is necessary in relation to the} \hlblue{purposes} for which they are processed.} \\
        \midrule
        \rowcolor{gray!40}
        \textbf{California Privacy Rights Act (CPRA), USA} \hfill \href{https://cppa.ca.gov/}{cppa.ca.gov/} \\
        \midrule
        {\sl Section 1798.100 (a)(1) \& (a)(2)}: 
            \textit{[\ldots] A business shall \hlblue{not collect additional categories of (sensitive)} \hlblue{personal information} or use (sensitive) personal information collected for additional purposes that are incompatible with the \hlgreen{disclosed purpose for which the (sensitive) personal information was collected} without providing the consumer with notice consistent with this section.} \\
        \midrule
        \rowcolor{gray!40}
        \textbf{General Personal Data Protection Law (LGPD), Brazil} \hfill \href{https://lgpd-brazil.info/}{lgpd-brazil.info/} \\
        \midrule
        {\sl Article 6:} 
            \textit{Activities of processing of personal data shall be subject to the following principles,}\\
            {\sl I:} \textit{processing done for \hlgreen{legitimate, specific and explicit purposes of which the data subject is informed, with} \hlgreen{no possibility of subsequent processing that is incompatible with these purposes};}\\
            {\sl II:} \textit{\hlgreen{compatibility of the processing with the purposes communicated to the data subject}, in accordance with the context of the processing;}\\
            {\sl III:} \textit{limitation of the processing to the \hlblue{minimum necessary to achieve its purposes}, covering data that are \hlblue{relevant, proportional and non-excessive in relation to purposes of the data processing} [\ldots];}\\
        \bottomrule
    \end{tabular}
    \vspace{4pt}
    \caption{Excerpts from various data protection regulations on the data minimization principle, highlighting \hlgreen{purpose limitation} and \hlblue{data relevance.}}
    \label{tab:legal}
    \vspace{-2em}
\end{table}

\subsection{Implications in practice}
\label{sec:legal_implication}
Building on the principles of purpose limitation and data relevance, this section examines three practical implications of data minimization as framed in the reviewed regulations:
\textbf{(1) Individualized nature of data minimization:} 
Notice that different individuals may require different amounts and types of information to fulfill a given purpose. 
For instance, in a loan approval scenario, personal data such as age, income, and job history are crucial. However, the importance of additional data, like medical history, can vary significantly among applicants. This variability underscores that data relevance—-and therefore redundancy—-is not universal but contextually and individually defined.
\textbf{(2) Data minimization is data dependent:}
Regulations describe the term ``data collection'' as acquiring data from entities like government agencies, which typically involves selecting relevant data from an already gathered large dataset. For instance, census data is often used for various analysis tasks. This differs from \emph{field} data collection, which lacks flexibility for individualized minimization adjustments. Our framework focuses on the former interpretation, wherein minimization is contingent on the data itself (as shown by the two distinct stages of ``Data collection from individuals'' and ``Data minimization'' in Figure \ref{fig:framework}).
\textbf{(3) Privacy expectations through minimization:}
There is an implicit expectation of privacy through minimization in data protection regulations~\citep{leemann2022prefer,goldsteen2021data,staab2024principle}. 
{\em However, this expectation overlooks a crucial aspect of real-world data--the inherent correlations among various features}. Information about individuals is rarely isolated, 
thus, merely removing or not collecting data, may allow for confident reconstruction inference~\citep{garfinkel2019understanding}, as we will show in our work.

\section{A formal framework for data minimization in ML}
\label{sec:formalization}
This section presents a data minimization framework for ML that aligns with the regulatory objectives discussed above.   
Successively, we examine the tension between these minimization objectives and their privacy implication and define various threat models to quantify real-world privacy risks. 

\subsection{Data minimization as optimization}
Consider a dataset $D = \{\bX, \bY\} = \{(\bm{x}_i, y_i)\}^{n}_{i=1}$, drawn i.i.d.~from an unknown distribution $D \sim (\cX, \cY)$. Therein, $\bm{x}_i \in \bX$ is a $p$-dimensional feature vector and $y_i \in \bY$ is the output label. 
Given a non-negative \emph{loss function} $\ell: \cY \times \cY \to \RR_+$, the goal of learning is to learn a predictor $f_\theta : \cX \to \cY$, with real-valued parameters $\theta$ that minimizes the empirical risk function (i.e., ERM):
\begin{align}
\label{eq:erm}
\optimal{\theta} \,= \argmin_\theta\; J(\theta; \bX, \bY) \quad \quad
\textrm{where} \quad J(\theta; \bX, \bY) = 
\frac{1}{n} \sum_{i=1}^n \ell(f_\theta(\bm{x}_i), y_i).
\end{align}

\emph{Data minimization} can be modelled as minimizing the size of the dataset $D$ by removing entries from the features $\bm{x}_i$ (\hlblue{data relevance}) while also retaining similar performance as it would have on the complete dataset (\hlgreen{purpose}). This objective can be formalized in the following bi-level optimization: 
\begin{subequations}
\label{eq:dm_goal}
    \begin{align}
        \minimize_{\bm{B} \in \{\bot, 1\}^{n \times p}} \;\;
            & \mhlblue{\| \bm{B} \|_1}
        \quad \quad
        \textsl{s.t.}:\;\; 
            J(\mhlgreen{\hat{\theta}}; \bX, \bY) 
            - J(\optimal{\theta}; \bX, \bY)  \leq \alpha
            \label{eq:fidelity}\\
            & \mhlgreen{\hat{\theta} = 
                \argmin_\theta \frac{1}{n} \sum_{i=1}^n 
                \ell\left(f_\theta( \mhlblue{\bm{x}_i \odot \bm{B}_i}), y_i\right)}
                \label{eq:purpose_limitation}
    \end{align}
\end{subequations}
where $\bm{B}$ is an $n \times p$ binary matrix, which we call the \emph{minimization matrix} (visualized in Figure~\ref{fig:datamin}), taking values in the set $\{\bot, 1\}$, and the $\ell_1$-norm of $\bm{B}$, i.e., $\| \bm{B} \|_1$, is simply the count of $1$ in the minimization matrix. Here, $\bot$ represents the concealment or removal of a value, i.e., $\forall v \in \R: v \times \bot = \bot$, and $\alpha \geq 0$ is a parameter which bounds the permitted drop in model quality due to data minimization. While we define ERM as our learning objective and the difference in loss on the original dataset as the quality drop (\eqref{eq:fidelity}), these choices can be easily generalized. 


The minimized input features $\bX'$ are defined as the element-wise product of the original features $\bX$ and the minimization matrix $\bm{B}$, i.e., $\bX' = \bX \odot \bm{B}$. 
Note that, data minimization produces features in the space $\cX \cup \{\bot\}$. While some learning algorithms can handle missing values ($\bot$), data imputation is needed in other cases. A discussion of imputation methods is delegated to \S \ref{sec:experiment_setup}.

The optimization problem in \eqref{eq:dm_goal} defines an operational method to remove entries from the feature set $\bX$ in a \emph{individualized} manner,
while adhering to pre-specified quality drop requirements on the original dataset (\eqref{eq:fidelity}), for the final model trained on the minimized dataset (\eqref{eq:purpose_limitation}). While this optimization captures the legal formulation of data minimization, it is however intractable to solve in practice. A discussion of tractable approximate algorithmic alternatives is delegated to \S \ref{sec:min_utility} and we discuss next the privacy implications of this data minimization process.

\subsection{Privacy leakage}
\label{sec:threat_models}

The implicit objective of data minimization is to enhance privacy. Thus, the paper focuses on 
assessing the minimized data using several threat models (Figure \ref{fig:framework}), to measure inference attacks under various objectives and degrees of access available to the adversary. 

\emph{Note that the expectation of data privacy from minimization in various regulations is focused on the events of a data breach}, i.e., situations in which the adversary has access to the minimized dataset. For this reason, we primarily focus on reconstruction and re-identification risks. 
This is different from conventional private ML settings~\citep{dwork2006differential,rigaki2023survey}, which focuses on information leakage due to ML inference (i.e., outside the "wall" in Figure \ref{fig:framework}).  Nonetheless, we also examine how data minimization privacy promise may hold under membership inference attacks and delegate this analysis to \S \ref{app:mia}.



\noindent\textbf{Reconstruction Risk (RCR).}
Real-world datasets often exhibit underlying associations between various features, making it possible to reconstruct minimized data~\citep{garfinkel2019understanding}. Reconstruction attacks thus aim to recover missing information from a target dataset.
Given the minimized dataset $\bX'$, the attacker's goal is to generate their reconstruction $\bX^R$ of the original set of features $\bX$. The ReConstruction Risk (RCR) can be evaluated by measuring the similarity between the original features $\bm{x}_i \in \bX$ and the reconstructed features $\bm{x}_i^R \in \bX^R$, computed using a gaussian kernel with $\sigma=1$~\citep{srebro2007good} as:
\begin{align}
    \textrm{RCR} = \frac{1}{n} \sum\limits_{i=1}^{n} e^{-|| \bm{x}_i - \bm{x}_i^R ||_2}.
\end{align}
If required, the reconstruction risk metric can also be adjusted further to prioritize the reconstruction of certain features over others by introducing appropriate weights to the similarity measurement.

\noindent\textbf{Re-identification Risk (RIR).}
%
Data breaches often lead to re-identification of individuals using partial or anonymized data matched with an auxiliary dataset $\bX^A$. The success of these re-identification attacks can be measured by the mean reciprocal rank (MRR) scores. For each data point $\bm{x}_i^A$ in $\bX^A$, the adversary ranks data points in the minimized dataset $\bX'$ based on the likelihood of matching identities. The Re-Identification Risk (RIR) is calculated as the average MRR score, assigning a score of $1$ for a correct match and $0$ otherwise, defined by the formula:
\begin{align}
     \textrm{RIR} = \frac{1}{n} \sum\limits_{i=1}^{n} \frac{1}{\textrm{index}(i, \textrm{rank}(\bm{x}_i^A, \bX'))}.
\end{align}
In this, $\textrm{rank}(\bm{x}_i^A, \bX')$ is the adversary's predicted ranking, and $\textrm{index}(e, A)$ gives the index of element $e$ in array $A$. It assumes a one-to-one match between auxiliary and minimized datasets.

\section{Data minimization and utility}
\label{sec:min_utility}


Having defined data minimization for ML as a bilevel optimization problem, this section outlines how to operationalize data minimization, with three strong baselines and three algorithms from the bi-level optimization literature to address the challenge formulated in \eqref{eq:dm_goal}.
Note that although the data minimization principle aims to optimize dataset size while maintaining model quality (see Constraint \ref{eq:fidelity}),  
these algorithms adopt a dual approach, optimizing model quality when trained on minimized data under a given sparsity constraint $\|\bm{B}\|_1 \leq k$. 
This dual approach offers the advantage of allowing these algorithms to find a sparsity parameter $k$ that meets the desired $\alpha$-drop in performance.

\begin{figure}
    \centering
    \includegraphics[width=1\linewidth]{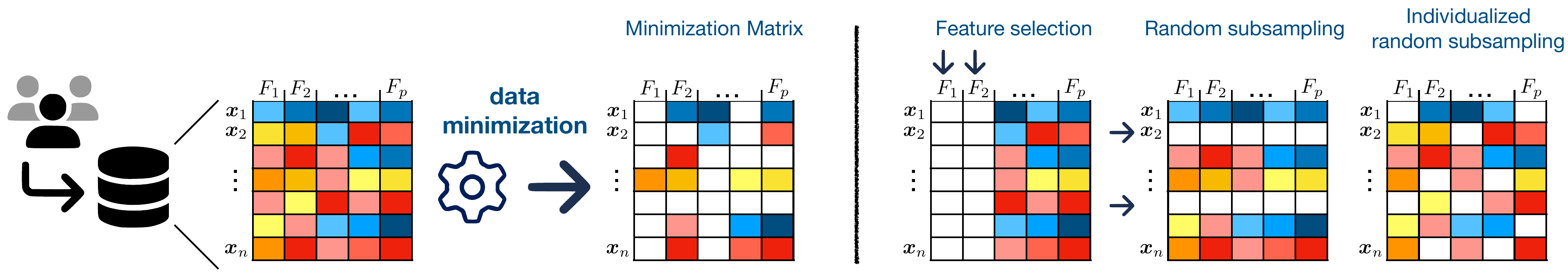}
    \caption{\textbf{(Left)} Data minimization as a set of individualized binary decisions, visualizing the minimization matrix $\bm{B}$. \textbf{(Right)} Three baseline algorithms for data minimization. 
    }
    \label{fig:datamin}
    \vspace{-1em}
\end{figure}

\noindent\textbf{Baseline techniques.}
The experiments implement three baseline methods, depicted pictorially in Figure \ref{fig:datamin}: (1) feature selection, (2) random subsampling, and (3) individualized random subsampling. {\em Feature selection}~\citep{blum1997selection} employs a breadth-based strategy that identifies and minimizes less important features within the dataset. {\em Random subsampling} is a depth-based strategy that randomly selects a subset of data points, thereby reducing the dataset size by excluding specific rows. {\em Individualized random subsampling} further tailors this approach by randomly selecting specific entries (feature, sample) for each individual, aiming to reduce dataset size in a more personalized manner. While these baselines help to assess model quality degradation, they do not fully comply with legal standards of data minimization, which require adherence to principles of purpose and data relevance, as discussed earlier and further expanded in \S \ref{sec:app_baseline}. 




\noindent\textbf{Data minimization algorithms.}
We briefly introduce three classes of algorithms adopted to solve the optimization posed in Problem~\ref{eq:dm_goal}: 
\begin{enumerate}[leftmargin=*, parsep=0pt, itemsep=0pt, topsep=0pt]
\item  {\em Approximating the target utility} \citep{6079692}, which focuses on finding an approximate closed-form solution to the lower-level problem, thus simplifying the original optimization. 

\item {\em Modelling the target utility} \citep{wang2006review}, which instead attempts to model the lower-level mapping and estimate it online again without solving the optimization. 

\item {\em Evolutionary algorithms} \citep{sinha2014finding}, which trade the advantage of carrying no assumption with slow convergence and high computational demand. 
\end{enumerate}
Extensive details about these algorithms, their theoretical underpinning, and comparative analysis can be found in \S \ref{sec:app_algorithms}.
Throughout our experimental evaluation, baseline methods are depicted with blue colours while the optimization algorithms for data minimization with red colours. 

\subsection{Experiment setup}
\label{sec:experiment_setup}

\noindent\textbf{Datasets.} Our evaluations focus on classification for both tabular data (marked with symbol $^\diamond$) and image data (symbol \(^\heartsuit\)), to cover a diverse set of distributions and modalities. In this section, we use \textbf{(i) the bank marketing dataset}~\citep{moro2014data},
taken from a telemarketing campaign with financial information of 11,162 customers,
and (\textbf{ii) the handwritten digits dataset}~\citep{xu1992methods}, containing a total of 1,797 handwritten digit images across 10 different classes (i.e., digits), 
Results on several additional datasets and modalities, including text, are reported in \S \ref{sec:app_defense}.


\noindent\textbf{Data splits.} Datasets are split into two equal groups, private and public data. 
The public data is used as the test data, to calculate statistics for data imputation, and train reference models for membership inference. The private data is further split into two equal groups, i.e. members and non-members. The members are our training data $\bX$, i.e., also the data that will be minimized in our pipeline. The non-members are only used for evaluating membership inference attacks (details in \S \ref{app:mia}).


\noindent\textbf{Learning setup and data imputation.} 
The experiments use logistic regression for the tabular dataset, and a fully connected neural network 
for the handwritten digits dataset. Both datasets are normalized using MinMaxScaler. Data imputation is performed under the assumption of a multivariate Gaussian distribution, utilizing the mean and covariance matrix of the public data to fill missing values with the mean of the conditional distribution~\citep{hoff2007extending}.

\subsection{Efficacy of data minimization}

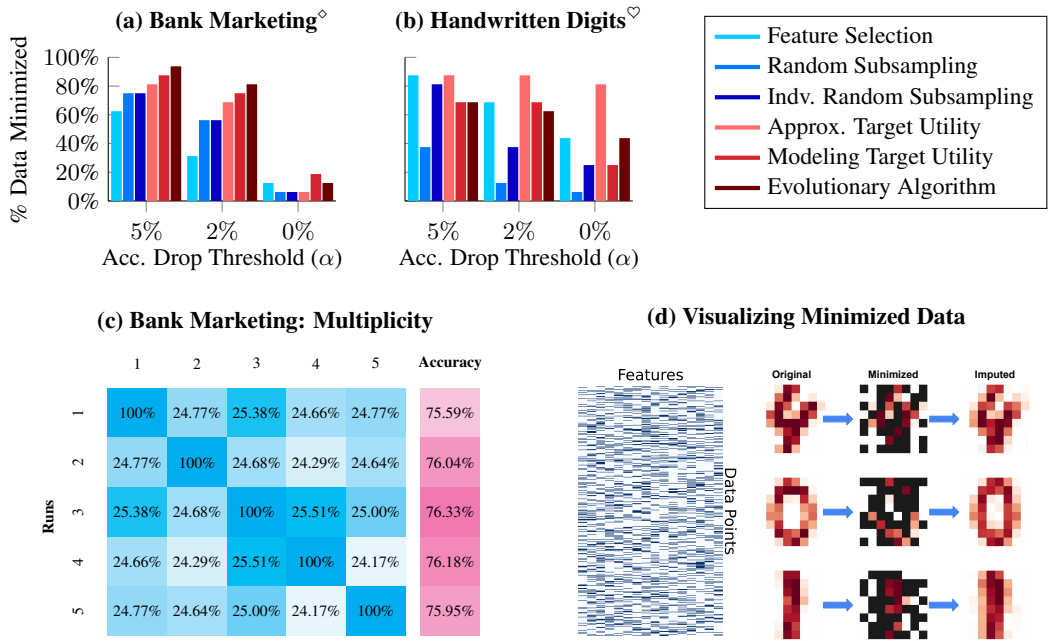
\begin{figure}[tb!]
    \captionsetup[subfigure]{labelformat=empty}
    \centering
    \footnotesize
    \begin{tikzpicture}
\begin{axis}[width=0.33\linewidth,height=0.25\linewidth,yticklabel= {\pgfmathprintnumber\tick\%},ylabel={\% Data Minimized},ymin=0,ymax=100,xlabel={Acc. Drop Threshold ($\alpha$)}, enlarge x limits=0.25,xtick=data,ybar=0.5pt,bar width=4pt,axis y line*=left,axis x line*=bottom,symbolic x coords={{$5\%$},{$2\%$},{$0\%$}},ylabel near ticks,x tick label style  = {text width=2cm,align=center},y label style  = {text width=4cm,align=center}, title={\textbf{(a) Bank Marketing$^\diamond$}}
]

\addplot[style={fill=featsel},draw=none
]
coordinates {({$0\%$},12.50)
             ({$2\%$},31.25)
	 	({$5\%$},62.50)
	     };
\addplot[style={fill=randsub},draw=none
]
coordinates {({$0\%$},6.25)
             ({$2\%$},56.25)
	 	({$5\%$},75.00)
	     };
\addplot[style={fill=perrandsub},draw=none
]
coordinates {({$0\%$},6.25)
             ({$2\%$},56.25)
	 	({$5\%$},75.00)
	     };
\addplot[style={fill=taylor},draw=none
]
coordinates {({$0\%$},6.25)
             ({$2\%$},68.75)
	 	({$5\%$},81.25)
	     };
\addplot[style={fill=metamodel},draw=none
]
coordinates {({$0\%$},18.75)
             ({$2\%$},75.00)
	 	({$5\%$},87.50)
	     };
\addplot[style={fill=genetic},draw=none
]
coordinates {({$0\%$},12.50)
             ({$2\%$},81.25)
	 	({$5\%$},93.75)
	     };

\end{axis}
\end{tikzpicture}%
\begin{tikzpicture}[trim left=-1em]
\begin{axis}[width=0.33\linewidth,height=0.25\linewidth,yticklabel= {\empty},ylabel={\color{white}\% Data Minimized},ymin=0,ymax=100,xlabel={Acc. Drop Threshold ($\alpha$)}, enlarge x limits=0.25,xtick=data,ybar=0.5pt,bar width=4pt,axis y line*=left,axis x line*=bottom,symbolic x coords={{$5\%$},{$2\%$},{$0\%$}},ylabel near ticks,x tick label style  = {text width=2cm,align=center},y label style  = {text width=4cm,align=center}, title={\textbf{(b) Handwritten Digits$^\heartsuit$}}
]

\addplot[style={fill=featsel},draw=none
]
coordinates {({$0\%$},43.75)
             ({$2\%$},68.75)
	 	({$5\%$},87.50)
	     };
\addplot[style={fill=randsub},draw=none
]
coordinates {({$0\%$},6.25)
             ({$2\%$},12.50)
	 	({$5\%$},37.50)
	     };
\addplot[style={fill=perrandsub},draw=none
]
coordinates {({$0\%$},25.00)
             ({$2\%$},37.50)
	 	({$5\%$},81.25)
	     };
\addplot[style={fill=taylor},draw=none
]
coordinates {({$0\%$},81.25)
             ({$2\%$},87.50)
	 	({$5\%$},87.50)
	     };
\addplot[style={fill=metamodel},draw=none
]
coordinates {({$0\%$},25.00)
             ({$2\%$},68.75)
	 	({$5\%$},68.75)
	     };
\addplot[style={fill=genetic},draw=none
]
coordinates {({$0\%$},43.75)
             ({$2\%$},62.50)
	 	({$5\%$},68.75)
	     };

\end{axis}
\end{tikzpicture}%
\hfill
\raisebox{3.3em}{
\begin{tikzpicture}[trim left=-3em]
\begin{axis}[scale=0.01,
legend cell align={left},
hide axis,
xmin=0, xmax=1,
ymin=0, ymax=1,
legend columns=1,
legend style={font=\footnotesize,/tikz/every even column/.append style={column sep=0.1cm}}
]
\addlegendimage{ultra thick, featsel}
\addlegendentry{Feature Selection};
\addlegendimage{ultra thick, randsub}
\addlegendentry{Random Subsampling};
\addlegendimage{ultra thick, perrandsub}
\addlegendentry{Indv. Random Subsampling};
\addlegendimage{ultra thick, taylor}
\addlegendentry{Approx. Target Utility};
\addlegendimage{ultra thick, metamodel}
\addlegendentry{Modeling Target Utility};
\addlegendimage{ultra thick, genetic}
\addlegendentry{Evolutionary Algorithm};

\end{axis}
\end{tikzpicture}
}
    \begin{subfigure}[b]{0.5\textwidth}
        \centering
        \caption{\textbf{(c) Bank Marketing: Multiplicity}}
        \begin{tikzpicture}[inner sep=0., scale=0.66, transform shape]
    \foreach \y [count=\n] in {{100, 24.77, 25.38, 24.66, 24.77}, {24.77, 100, 24.68, 24.29, 24.64}, {25.38, 24.68, 100, 25.51, 25.00}, {24.66, 24.29, 25.51, 100, 24.17}, {24.77, 24.64, 25.00, 24.17, 100}}{
        \foreach \x [count=\m] in \y {
            \pgfmathsetmacro \clr {(\x-24)*50}
            \pgfmathsetmacro \ns {\n*1.0}
            \pgfmathsetmacro \ms {\m*1.2}
            \node[fill=cyan!\clr, minimum width=12mm, minimum height=10mm] at (\ms,-\ns) {\small\x\%};
        }
    }

    \node[rotate=90] at (-0.6, -3.1) {\textbf{Runs}};
    \foreach \xlabel [count=\m] in {1,2,3,4,5}{
        \pgfmathsetmacro \ms {\m*1.2};
        \node[minimum width=12mm, minimum height=10mm] at (\ms,0) {\xlabel};
        \node[minimum width=12mm, minimum height=10mm,rotate=90] at (0,-\m) {\xlabel};
    }

    \foreach \x [count=\n] in {75.59, 76.04, 76.33, 76.18, 75.95} {
        \pgfmathsetmacro \clr {(\x-75)*50}
        \pgfmathsetmacro \ns {\n*1.0}
        \node[fill=magenta!\clr, minimum width=12mm, minimum height=10mm] at (7.5,-\ns) {\small\x\%};
    }

    \node[] at (7.5,0.0) {\textbf{Accuracy}};
\end{tikzpicture}
    \end{subfigure}
    \hfill
    \begin{subfigure}[b]{0.47\textwidth}
        \centering
        \caption{\textbf{(d) Visualizing Minimized Data}}
        \vspace{1.2em}
        \begin{subfigure}[b]{0.34\textwidth}
             \centering
             \includegraphics[width=\textwidth]{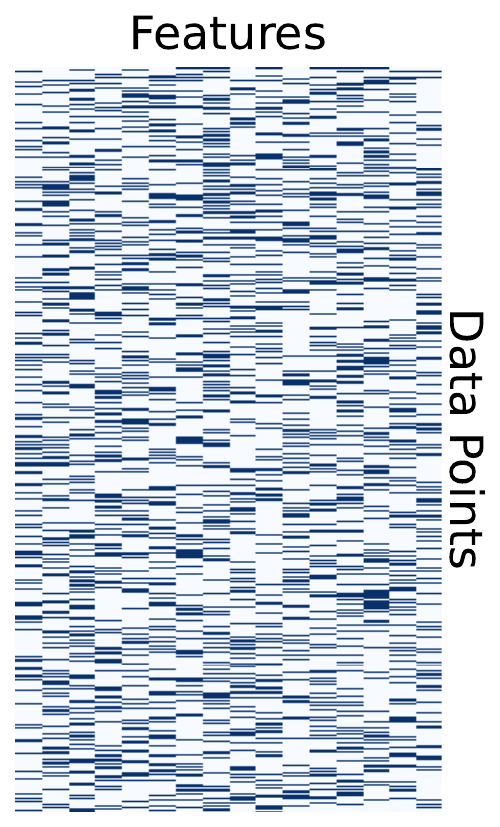}
        \end{subfigure}
        \begin{subfigure}[b]{0.59\textwidth}
             \centering
             \includegraphics[width=\textwidth]{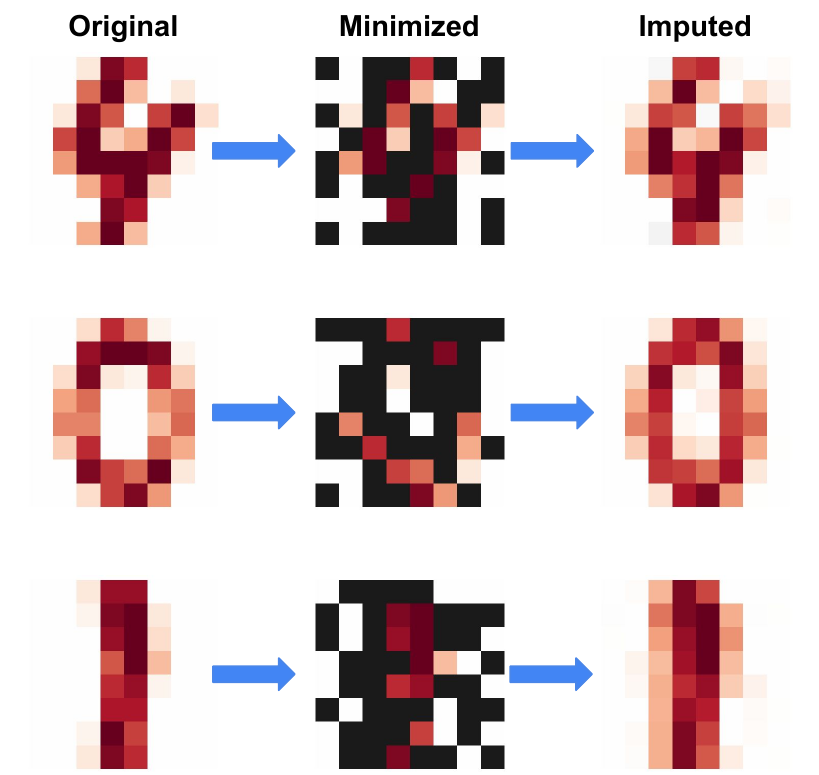}
        \end{subfigure}
    \end{subfigure}
    \caption{\textbf{(a, b)} Percentage of dataset minimized under different accuracy drop thresholds ($\alpha$). 
    \textbf{(c)} Pairwise overlap between minimized datasets across five runs with changing randomness. 
    \textbf{(d)} Visualizing minimized datasets across modalities showing: (Left) emergent individualized minimization in a compressed view of minimized bank dataset; and (Right) misalignment of minimization with privacy as imputed images have reconstructed missing information despite minimization.}
    \label{fig:dataset_size}
    \vspace{-1.5em}
\end{figure}

First, this section assesses the ability of data minimization algorithms to reduce the dataset size across several modalities. Figure \ref{fig:dataset_size} summarizes the results, comparing the amount of data that could be minimized (y-axis) given a maximum drop in accuracy (x-axis) when compared to the accuracy returned by the classifier trained over the original, not-minimized, dataset. 
The results indicate a strong trend: {\em A substantial amount of data can often be removed without sacrificing utility, suggesting that much of the collected data is superfluous in the datasets analyzed.} 

Notably, the baseline methods for minimization are found effective in reducing substantial amounts of data, possibly because they leverage existing structural redundancies in the dataset. For instance, feature selection excels with the handwritten digit dataset, where outer boundary pixels are never covered by the digits, and thus can be easily removed with minimal impact on model performance. 
Nonetheless, the optimization algorithms consistently outperform the baselines across all datasets, with different algorithms outshining others in specific scenarios. Evolutionary algorithms, for instance, surpass others on the bank dataset, driving minimization to extreme sparsity. It highlights a strength of minimization at its core: reducing the dataset size to a mere $18.75\%$ (and $6.25\%$) subset of the original bank dataset still allows the model to maintain utility with only a $2\%$ (and $5\%$) drop in accuracy. Similarly, approximating the target utility excels on the handwritten digits dataset, presumably due to its compatibility with the values of redundant features in the dataset (empty pixels are represented with $0$, compatible with the zero-imputation assumption of the algorithm). 
An in-depth discussion of the strengths and weaknesses of these algorithms is provided in \S \ref{sec:app_algorithms}.


\subsection{Multiplicity, emergent individualization, and privacy in data minimization}
\label{sec:multiplicity}

Building on the previous section's demonstration of data minimization's effectiveness in various settings, this section explores \emph{the concept of multiplicity of data minimization}. We applied the evolutionary algorithm five times to the bank dataset under different randomness settings, each targeting 75\% sparsity. 
Notably, the results, illustrated in Figure~\ref{fig:dataset_size}(c), show minimal overlap among the datasets retained from each run, with the highest overlap being only 25.51\%.
Clearly, there are many distinct ways to achieve data minimization while maintaining utility that meets data relevance constraints (i.e., no further data can be removed without compromising utility).
Such variability underscores the flexibility of achieving data minimization through diverse algorithms. However, this also suggests that different minimized datasets may pose varying privacy risks related to the undisclosed features, as discussed in \S \ref{sec:legal_implication}, pointing to a misalignment in privacy outcomes.



Next, we visualize the minimized dataset and highlight some intriguing characteristics. Starting with the bank dataset, minimized to 75\% sparsity using the evolutionary algorithm, Figure \ref{fig:dataset_size}(d)(Left) shows that individualization trends naturally emerge from the optimization process, where no single feature is consistently favoured over others.  
Similarly, the minimized images from the handwritten digits dataset, processed by the evolutionary algorithm at 50\% sparsity, are shown in Figure \ref{fig:dataset_size}(d)(Right). The trends here are more interpretable; for example, the central vertical line is preserved in the image of the digit `1', while the outer curves are retained for `0'.
Notice however that despite reducing the dataset to half its original size, a significant portion of the images can still be reconstructed using statistics learned from public data.
This provides a strong indication of privacy risks and suggests that, as we will show next, a minimized dataset does not equate to enhanced privacy.

\section{Data minimization and privacy}
\label{sec:min_privacy}
In the previous section, we showed that various algorithms can effectively minimize data while focusing on utility. However, as discussed in \S \ref{sec:legal_implication}, there is an expectation of privacy associated with minimization, currently unexplored in literature~\citep{rastegarpanah2021auditing,shanmugam2022learning,biega2020operationalizing}. Next, we conduct an important empirical assessment of minimization algorithms on real-world privacy attacks and their alignment with the minimization objective.

\noindent\textbf{Defining inference attacks.} During a reconstruction attack, the attacker aims to recover missing information from the minimized dataset. To achieve this, we use the data imputation method described in \S \ref{sec:experiment_setup}. 
In contrast, a re-identification attack involves the attacker attempting to identify an individual using only partial information. Specifically, the attacker aims to find the best match in the minimized dataset $\bX'$ for a data point from the auxiliary dataset (in our setup, $\bX^A \equiv \bX$). For this purpose, we use a modified Euclidean distance that disregards missing values and adjusts the distance scaling accordingly, to accurately rank the best matches.





\subsection{Evaluating privacy leakage}
\label{sec:privacy_leackage}




\begin{figure*}
    \centering
    \footnotesize
    \begin{tikzpicture}
\begin{axis} [width=0.35\linewidth,height=0.25\linewidth,yticklabel= {\pgfmathprintnumber\tick\%},ymin=0,ymax=100,xticklabel= {\pgfmathprintnumber\tick},xmin=0,xmax=1,ylabel={Dataset Size},xlabel={Re-identification Risk{\color{white}p}},axis y line*=left,axis x line*=bottom,ylabel near ticks,ylabel style={align=center, text width=4cm},xtick={0,0.5,1},legend pos=south east, legend style={font=\tiny}, legend cell align={left}, title={\textbf{(a) Bank Marketing$^\diamond$}}]

\addplot [featsel,no marks,ultra thick] table[x=Reid,y=Size] {data/bank/feature_selection.dat};
\addplot [randsub,no marks,ultra thick] table[x=Reid,y=Size] {data/bank/random_subsampling.dat};
\addplot [perrandsub,no marks,ultra thick] table[x=Reid,y=Size] {data/bank/personalized_random_subsampling.dat};
\addplot [taylor,no marks,ultra thick] table[x=Reid,y=Size] {data/bank/taylor_approx_heuristic.dat};
\addplot [genetic,no marks,ultra thick] table[x=Reid,y=Size] {data/bank/genetic_heuristic.dat};
\addplot [metamodel,no marks,ultra thick] table[x=Reid,y=Size] {data/bank/metamodeling_heuristic.dat};

\end{axis}
\end{tikzpicture}%
\begin{tikzpicture}
\begin{axis} [width=0.35\linewidth,height=0.25\linewidth,yticklabel= {\empty},ymin=0,ymax=100,xticklabel= {\pgfmathprintnumber\tick},xmin=0,xmax=1,ylabel={\color{white}Dataset Size},xlabel={Re-identification Risk{\color{white}p}},axis y line*=left,axis x line*=bottom,ylabel near ticks,ylabel style={align=center, text width=4cm},xtick={0,0.5,1},legend pos=south east, legend style={font=\tiny}, legend cell align={left}, title={\textbf{(b) Handwritten Digits$^\heartsuit$}}]

\addplot [featsel,no marks,ultra thick] table[x=Reid,y=Size] {data/digits_nn/feature_selection.dat};
\addplot [randsub,no marks,ultra thick] table[x=Reid,y=Size] {data/digits_nn/random_subsampling.dat};
\addplot [perrandsub,no marks,ultra thick] table[x=Reid,y=Size] {data/digits_nn/personalized_random_subsampling.dat};
\addplot [taylor,no marks,ultra thick] table[x=Reid,y=Size] {data/digits_nn/taylor_approx_heuristic.dat};
\addplot [genetic,no marks,ultra thick] table[x=Reid,y=Size] {data/digits_nn/genetic_heuristic.dat};
\addplot [metamodel,no marks,ultra thick] table[x=Reid,y=Size] {data/digits_nn/metamodeling_heuristic.dat};

\end{axis}
\end{tikzpicture}%
\hfill
\begin{tikzpicture}
\begin{axis}[width=0.28\linewidth,height=0.25\linewidth,,ylabel={Re-identification Risk \\($\alpha=5\%$)},ymin=0,ymax=1,xlabel={Privacy Coeff. ($\beta$)}, enlarge x limits=0.25,xtick=data,ybar=0.5pt,bar width=4pt,axis y line*=left,axis x line*=bottom,symbolic x coords={{$0$},{$1.5$},{$3$}},ylabel near ticks,x tick label style  = {text width=2cm,align=center},y label style  = {text width=4cm,align=center},title={\textbf{(c) Handwritten Digits$^\heartsuit$}}
]

\addplot[style={fill=taylor},draw=none
]
coordinates {({$0$},0.41)
             ({$1.5$},0.19)
	 	({$3$},0.41)
	     };
\addplot[style={fill=metamodel},draw=none
]
coordinates {({$0$},1)
             ({$1.5$},0.3)
	 	({$3$},0.45)
	     };
\addplot[style={fill=genetic},draw=none
]
coordinates {({$0$},1)
             ({$1.5$},0.26)
	 	({$3$},0.53)
	     };

\end{axis}
\end{tikzpicture}%
    \begin{tikzpicture}
\begin{axis} [width=0.35\linewidth,height=0.25\linewidth,yticklabel= {\pgfmathprintnumber\tick\%},ymin=0,ymax=100,xticklabel= {\pgfmathprintnumber\tick},xmin=0,xmax=1,ylabel={Dataset Size},xlabel={Reconstruction Risk},axis y line*=left,axis x line*=bottom,ylabel near ticks,ylabel style={align=center, text width=4cm},xtick={0,0.5,1},legend pos=south east, legend style={font=\tiny}, legend cell align={left}, title={\textbf{(d) Bank Marketing$^\diamond$}}]

\addplot [featsel,no marks,ultra thick] table[x=Recon,y=Size] {data/bank/feature_selection.dat};
\addplot [randsub,no marks,ultra thick] table[x=Recon,y=Size] {data/bank/random_subsampling.dat};
\addplot [perrandsub,no marks,ultra thick] table[x=Recon,y=Size] {data/bank/personalized_random_subsampling.dat};
\addplot [taylor,no marks,ultra thick] table[x=Recon,y=Size] {data/bank/taylor_approx_heuristic.dat};
\addplot [genetic,no marks,ultra thick] table[x=Recon,y=Size] {data/bank/genetic_heuristic.dat};
\addplot [metamodel,no marks,ultra thick] table[x=Recon,y=Size] {data/bank/metamodeling_heuristic.dat};

\end{axis}
\end{tikzpicture}%
\begin{tikzpicture}
\begin{axis} [width=0.35\linewidth,height=0.25\linewidth,yticklabel= {\empty},ymin=0,ymax=100,ylabel={\color{white}Dataset Size},xticklabel= {\pgfmathprintnumber\tick},xmin=0,xmax=1,xlabel={Reconstruction Risk},axis y line*=left,axis x line*=bottom,ylabel near ticks,ylabel style={align=center, text width=4cm},xtick={0,0.5,1},legend pos=south east, legend style={font=\tiny}, legend cell align={left}, title={\textbf{(e) Handwritten Digits$^\heartsuit$}}]

\addplot [featsel,no marks,ultra thick] table[x=Recon,y=Size] {data/digits_nn/feature_selection.dat};
\addplot [randsub,no marks,ultra thick] table[x=Recon,y=Size] {data/digits_nn/random_subsampling.dat};
\addplot [perrandsub,no marks,ultra thick] table[x=Recon,y=Size] {data/digits_nn/personalized_random_subsampling.dat};
\addplot [taylor,no marks,ultra thick] table[x=Recon,y=Size] {data/digits_nn/taylor_approx_heuristic.dat};
\addplot [genetic,no marks,ultra thick] table[x=Recon,y=Size] {data/digits_nn/genetic_heuristic.dat};
\addplot [metamodel,no marks,ultra thick] table[x=Recon,y=Size] {data/digits_nn/metamodeling_heuristic.dat};

\end{axis}
\end{tikzpicture}%
\hfill
\begin{tikzpicture}
\begin{axis}[width=0.28\linewidth,height=0.25\linewidth,,ylabel={Reconstruction Risk \\($\alpha=5\%$)},ymin=0,ymax=0.4,xlabel={Privacy Coeff. ($\beta$)}, enlarge x limits=0.25,xtick=data,ybar=0.5pt,bar width=4pt,axis y line*=left,axis x line*=bottom,symbolic x coords={{$0$},{$1.5$},{$3$}},ylabel near ticks,x tick label style  = {text width=2cm,align=center},y label style  = {text width=4cm,align=center},title={\textbf{(f) Handwritten Digits$^\heartsuit$}}
]

\addplot[style={fill=taylor},draw=none
]
coordinates {({$0$},0.24)
             ({$1.5$},0.09)
	 	({$3$},0.21)
	     };
\addplot[style={fill=metamodel},draw=none
]
coordinates {({$0$},0.29)
             ({$1.5$},0.22)
	 	({$3$},0.25)
	     };
\addplot[style={fill=genetic},draw=none
]
coordinates {({$0$},0.3)
             ({$1.5$},0.26)
	 	({$3$},0.36)
	     };

\end{axis}
\end{tikzpicture}
    \begin{tikzpicture}
\begin{axis}[scale=0.01,
legend cell align={left},
hide axis,
xmin=0, xmax=1,
ymin=0, ymax=1,
legend columns=3,
legend style={font=\footnotesize,/tikz/every even column/.append style={column sep=0.1cm}}
]
\addlegendimage{ultra thick, featsel}
\addlegendentry{Feature Selection};
\addlegendimage{ultra thick, randsub}
\addlegendentry{Random Subsampling};
\addlegendimage{ultra thick, perrandsub}
\addlegendentry{Indv. Random Subsampling};
\addlegendimage{ultra thick, taylor}
\addlegendentry{Approx. Target Utility};
\addlegendimage{ultra thick, metamodel}
\addlegendentry{Modeling Target Utility};
\addlegendimage{ultra thick, genetic}
\addlegendentry{Evolutionary Algorithm};

\end{axis}
\end{tikzpicture}
    \caption{\textbf{(a, b, d, e)} Re-identification and reconstruction risks under changing sparsity. While the overall trends show improvement in privacy with data minimization, they are not aligned with the trends of dataset size.
    \textbf{(c, f)} Privacy risks under feature-level privacy scores. The lowest risk values can be seen at $\beta=1.5$, highlighting the importance of considering privacy during minimization.}
    \label{fig:reid_recon}
    \vspace{-2em}
\end{figure*}
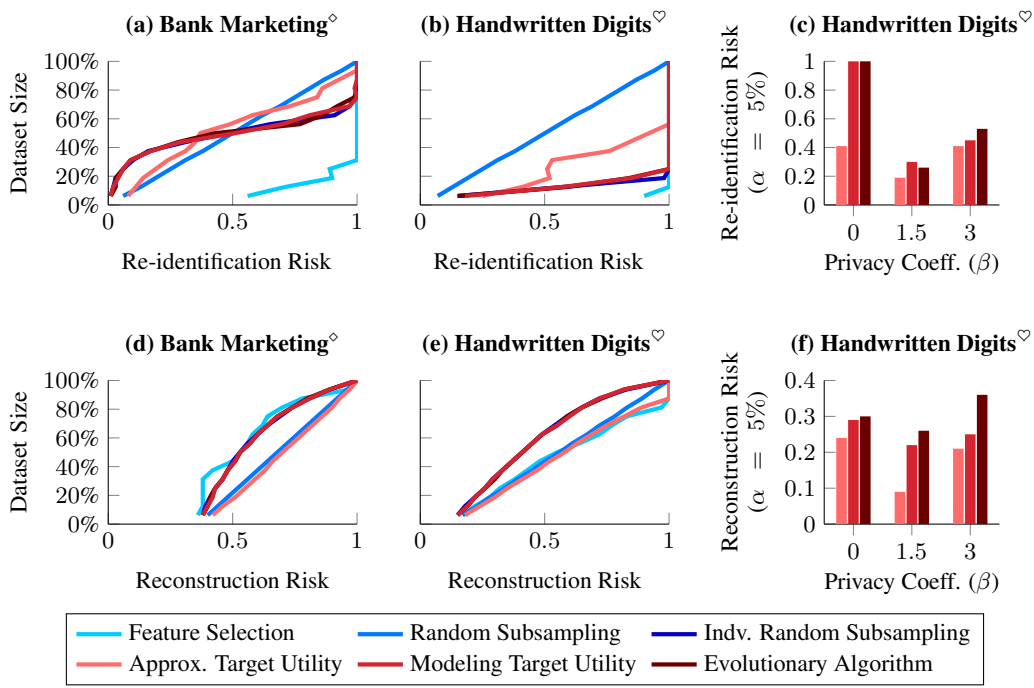

\noindent\textbf{Data minimization and re-identification risk.}
We first focus on the re-identification risks for the handwritten digits dataset, shown in Figure \ref{fig:reid_recon}(b). A key observation is that re-identification risk remains remarkably high, even after extensive minimization. For instance, the re-identification risk is close to $1$ for most algorithms even when the dataset is reduced to approximately 20\% of its original size. This behaviour is due to the dataset's large feature space. Thus, while minimization algorithms were able to reduce dataset size significantly while maintaining utility (as shown in Figure \ref{fig:dataset_size}), they do not achieve comparable reductions in re-identification risk. {\em This underscores a fundamental misalignment between the goals of data minimization and the actual privacy goals.}

While this misalignment is less pronounced in the bank dataset (Figure \ref{fig:reid_recon}(a)), there are still regions in which reducing dataset size does not correspondingly decrease re-identification risk. Notably, the algorithm that approximates the target utility demonstrates a closer alignment with privacy goals compared to other algorithms. Considering the lower amount of minimization by approximating the target utility as observed in Figure \ref{fig:dataset_size}, we can infer that the additional data minimized by the other two algorithms was not aligned with re-identification risk.
Thus, while these algorithms would have been preferred if we only considered utility and dataset size (as we did in Figure \ref{fig:dataset_size}), they do not provide proportionate privacy improvements, highlighting a critical concern of misalignment.

Finally, observe that our baseline algorithms display notable trends across both datasets. Random subsampling emerges as an effective strategy to reduce dataset size while being aligned with re-identification risks. Indeed, completely removing a data point also eliminates any risk of its re-identification.
In contrast, feature selection, despite preserving reasonable utility post-minimization, fails to effectively mitigate re-identification risks in any dataset.  This inadequacy can be attributed to the persistence of certain features in the minimized dataset that, even when isolated, can uniquely identify individuals. 
This phenomenon suggests a general misalignment between breadth-based minimization techniques~\citep{rastegarpanah2021auditing,goldsteen2021data,staab2024principle} and the expected reduction in re-identification risks. Once again, these observations highlight the disparity between mere reductions in dataset size and actual improvements in privacy outcomes.

\noindent\textbf{Data minimization and reconstruction risk.}
Next, we shift focus to the results of reconstruction risks,  shown in Figure~\ref{fig:reid_recon}(d, e). Notably, even at the highest levels of data minimization (smallest dataset size), reconstruction risks remain significant for both domains. 
This aspect is linked to the ability to reconstruct features using overall dataset statistics, which, although decreasing in accuracy as minimization increases, still retain some reconstructive value.  Interestingly, we find that the handwritten digits dataset is comparatively easier to protect against reconstruction, likely due to its higher feature variance relative to the bank dataset. While the same high variance in features led to worse re-identification risk, it instead helped the handwritten digits dataset achieve better alignment with reconstruction risk. Thus, not only is the dataset size not aligned with privacy, but even different privacy risks are not aligned with each other, motivating the explicit involvement of appropriate privacy constraints during minimization.

The trends observed in re-identification risk also manifest in reconstruction risks, albeit with some unique differences. Firstly, both random subsampling and individualized random subsampling have linear relationships between dataset size and privacy, highlighting a strong alignment with privacy. Secondly, the best-performing algorithm for dataset size might not always be the best choice overall. Specifically, while the algorithm that approximates the target utility shows strong performance on the bank marketing dataset and holds a significant advantage on the handwritten digits dataset, it consistently presents higher reconstruction risks for both datasets. 
These trends emphasize the importance of incorporating privacy considerations when selecting the most appropriate minimization algorithm for a specific scenario. 
\subsection{Adapting minimization for better privacy}
\label{sec:min_defenses}
We now present a simple yet effective adjustment to the minimization problem to align better with privacy and demonstrate the feasibility of a more comprehensive minimization objective, managing both utility and privacy.
Given the output of a minimization algorithm $\bm{B}^a = [\optimal{\bm{B}}]_{\bot \rightarrow 0}$, we define privacy scores $\bm{V}$, such that the score matrix $\bm{C}_{ij} = \bm{B}^a_{ij} + \beta\bm{V}_{ij}$ determines the indices that should be minimized. For simplicity, we will only define feature-level privacy scores, which are not personalized, i.e., $\bm{V}_{ij} = \bm{V}_{qj},\; \forall i, q \in [n], j \in [p]$.
Here, $\beta$ serves as a hyperparameter to tune the emphasis on privacy. Ultimately, we minimize values with the lowest $\bm{C}_{ij}$ scores to achieve the target sparsity. 
The privacy scores $\bm{V}$ are normalized to $[0, 1]$ before being combined with the minimization matrix.


\begin{itemize}[leftmargin=*, parsep=0pt, itemsep=0pt, topsep=0pt]
    \item \textbf{Privacy scores to reduce re-identification.} In re-identification attacks, the risk arises from disclosing \emph{unique} features (e.g., phone number, SSN) rather than non-unique features (e.g., gender, race), as they make it easier to identify individuals. Following this rationale, we propose using the negative of the number of unique values for a feature as its privacy score. Minimizing distinctive features can increase the difficulty of re-identification.
    \item \textbf{Privacy scores to reduce reconstruction.} In reconstruction attacks, the risk stems from features that exhibit a high correlation with other features, exploited by adversaries to deduce missing values. Thus, to mitigate reconstruction risks, we propose using the negative of the average correlation of each feature with every other feature as a measure of its independence and thus its privacy score.
\end{itemize}

\smallskip\noindent\textbf{Evaluating privacy score modifications.}
For a given level of sparsity, it can be expected that incorporating privacy scores will decrease data breach risks, but potentially reduce accuracy. The critical question is whether we can attain a more favourable trade-off between privacy and utility, regardless of the sparsity. We present the results in Figure \ref{fig:reid_recon}(c, f) for the handwritten digits dataset, varying the hyperparameter $\beta$. Results for other datasets are present in \S \ref{sec:app_defense}.

At $\beta=1.5$, integrating privacy scores enhances the privacy-utility trade-off, cutting the re-identification risk by more than half while maintaining the same accuracy drop. The impact on reconstruction risk is less dramatic, but improvements are evident. Increasing the focus on privacy to $\beta=3$ results in a less optimal trade-off, yet still better than at $\beta=0$. 
This notable effect of a basic feature-level privacy score underscores the necessity of directly addressing privacy in the minimization process, instead of relying on its incidental emergence.

\section{Discussion and conclusion}
\label{sec:discussion}
In proposing a formal framework for data minimization in ML, this paper reveals a disconnect between legal mandates and their practical implementation. While data protection regulations aim to limit data collection with an expectation of privacy, current objectives of minimization fall short of providing robust privacy safeguards. Notice, however, that this is not to say that minimization is incompatible with privacy;
instead, we emphasize the need for approaches that incorporate privacy into their objectives (as done in \S \ref{sec:min_defenses}), rather than treating them as an afterthought.


Addressing this concern also brings forth a variety of optimization challenges. 
The optimization problem is complicated further under alternative interpretations of data collection, such as gathering a range of data rather than exact values~\citep{goldsteen2021data,staab2024principle}.
These challenges are particularly significant in large-scale applications where both time and accuracy are critical factors. 
Therefore, future work should focus on the development of efficient minimization algorithms able to make good trade-offs between utility and privacy.
The ethical and fairness considerations of data minimization also add to its complexity. 
By design, data minimization is likely to remove data that resembles the majority~\citep{biega2020operationalizing}, leaving the minority more vulnerable to privacy risks. 
Thus, developing fairness-aware mechanisms for minimization is an important avenue for future research.

In summary, our study marks a crucial step in aligning the legal requirements with practical, technical solutions for data minimization in ML. 
It sets the stage for future work aimed at developing comprehensive, efficient, and ethically sound methodologies for minimization. 
It hopes to guide the creation of ML systems that are both respectful of individual privacy and compliant with legal norms, thereby emphasizing the need for multidisciplinary approaches to the challenges of data minimization.



\begin{ack}
This research is partially supported by NSF grants 2133169, 2232054 and NSF CAREER Award 2143706. Fioretto is also supported by an Amazon Research Award and a Google Research Scholar Award. Its views and conclusions are those of the authors only.
\end{ack}

\bibliographystyle{abbrvnat}
\bibliography{references}

\appendix
\setcounter{theorem}{0}
\setcounter{lemma}{0}
\setcounter{proposition}{0}

\section{Legal language}
\label{app:legal}

Table \ref{tab:legal_long} summarizes relevant excerpts from six data protection regulations from around the world. Along with \hlgreen{purpose limitation} and \hlblue{data relevance}, we also highlight excerpts that show the expectation of \hlred{data privacy} through minimization.


In various data protection regulations, data minimization is defined on personal data, emphasizing that it is the personal data that needs to be minimized. Personal data is defined as any information that directly \textit{identifies a particular individual}, or can be indirectly \textit{combined with other information to identify a particular individual} (PIPA, South Korea). It is clear that these regulations prioritize data minimization to avoid collecting unnecessary information that could compromise an individual's privacy, hence the expectation of enhanced privacy under minimization.

\begin{table*}[tbh!]
    \centering
    \footnotesize
    \begin{tabular}{p{0.95\linewidth}}
        \toprule
        \rowcolor{gray!40}
        \textbf{General Data Protection Regulation (GDPR), Europe} \hfill \href{https://gdpr-info.eu/}{gdpr-info.eu/} \\
        \midrule
        {\sl Article 4(1)}: 
            \textit{``personal data'' means any information relating to an \hlred{identified or identifiable natural person} (``data subject'') [\ldots];}
        \\[0.3em]
        {\sl Article 5(1)(b)}: 
            \textit{Personal data shall be collected for  \hlgreen{specified, explicit and legitimate purposes and not} \hlgreen{further processed in a manner that is incompatible with those purposes}\!;}\\[0.3em]
        {\sl Article 5(1)(c)}: 
            \textit{Personal data shall be \hlblue{adequate, relevant and limited to what is necessary in relation to the} \hlblue{purposes} for which they are processed.} \\
        \midrule
        \rowcolor{gray!40}
        \textbf{California Privacy Rights Act (CPRA), USA} \hfill \href{https://cppa.ca.gov/}{cppa.ca.gov/} \\
        \midrule
        {\sl Section 1798.100 (a)(1) \& (a)(2)}: 
            \textit{[\ldots] A business shall \hlblue{not collect additional categories of (sensitive) personal information} or use (sensitive) personal information collected for additional purposes that are incompatible with the \hlgreen{disclosed purpose for which the (sensitive) personal information was collected} without providing the consumer with notice consistent with this section.} \\[0.3em]
        {\sl Section 1798.140 (v)(1):} 
            \textit{``Personal information'' means information that \hlred{identifies, relates to, describes, is reasonably capable of being associated with, or could reasonably be} 
            \hlred{linked, directly or indirectly, with a particular consumer or household}.} \\
        \midrule
        \rowcolor{gray!40}
        \textbf{General Personal Data Protection Law (LGPD), Brazil} \hfill \href{https://lgpd-brazil.info/}{lgpd-brazil.info/} \\
        \midrule
        {\sl Article 5(I):} 
            \textit{personal data: information regarding an \hlred{identified or identifiable natural person};} \\[0.3em]
        {\sl Article 6:} 
            \textit{Activities of processing of personal data shall be subject to the following principles,}\\
            {\sl I:} \textit{processing done for \hlgreen{legitimate, specific and explicit purposes of which the data subject is informed, with} \hlgreen{no possibility of subsequent processing that is incompatible with these purposes};}\\
            {\sl II:} \textit{\hlgreen{compatibility of the processing with the purposes communicated to the data subject}, in accordance with the context of the processing;}\\
            {\sl III:} \textit{limitation of the processing to the \hlblue{minimum necessary to achieve its purposes}, covering data that are \hlblue{relevant, proportional and non-excessive in} \hlblue{relation to purposes of the data processing} [\ldots];}\\
        \midrule
        \rowcolor{gray!40}
        \textbf{Protection of Personal Information Act (POPIA), South Africa} \hfill \href{https://popia.co.za/}{popia.co.za/} \\
        \midrule
        {\sl Section 1}: 
            \textit{‘‘personal information’’ means information relating to an \hlred{identifiable, living, natural person}, [\ldots]}\\[0.3em]
        {\sl Section 10}: 
            \textit{Personal information may only be processed if, given the purpose for which it is processed, it is \hlblue{adequate, relevant and not excessive}.}\\[0.3em]
        {\sl Section 13(1)}: 
            \textit{Personal information must be collected for a \hlgreen{specific, explicitly defined and lawful purpose} [\ldots]} \\
        \midrule
        \rowcolor{gray!40}
        \textbf{Consumer Data Rights (CDR), Australia} \hfill \href{https://www.legislation.gov.au/Details/F2023C00735}{www.legislation.gov.au/Details/F2023C00735} \\
        \midrule
        {\sl Rule 1.8(a)}: 
            \textit{An accredited person complies with the data minimisation principle if when making a consumer data request on behalf of a CDR consumer, it does not seek to collect: (i)  \hlblue{more CDR data than is reasonably needed}; [\ldots]} \\[0.3em]
        {\sl Rule 4.11(1)(a)}: 
            \textit{When asking a CDR consumer to give a consent, an accredited person must allow the CDR consumer to choose the types of CDR data to which the consent will apply by enabling the CDR consumer to actively select or otherwise clearly indicate: (i)..; and (ii) in the case of a use consent―\hlgreen{the specific uses of collected data to which they are consenting}; [\ldots]}\\
        \midrule
        \rowcolor{gray!40}
        \textbf{Personal Information Protection Act (PIPA), South Korea} \hfill \href{https://www.pipc.go.kr/eng/index.do}{www.pipc.go.kr/eng/index.do} \\
        \midrule
        \sl{Article 2(1)}: 
            \textit{The term ``personal information'' means any of the following information relating to a living individual: 
            (a) Information that \hlred{identifies a particular individual} [\ldots]; 
            (b) Information which, even if it by itself does not identify a particular individual, \hlred{may be easily combined with} \hlred{other information to identify a particular individual} [\ldots];} \\[0.3em]
        {\sl Article 3(1)}: 
            \textit{The personal information controller shall \hlgreen{specify explicitly the purposes for which personal} \hlgreen{information is processed}; and shall collect personal information \hlblue{lawfully and fairly to the minimum extent necessary for such purposes}.} \\
        \bottomrule
    \end{tabular}
    \caption{Excerpts from various data protection regulations from across the globe on the principle of data minimization, highlighting language on \hlgreen{purpose limitation}, \hlblue{data relevance}, and references to the expectations of \hlred{data privacy}.}
    \label{tab:legal_long}
\end{table*}

\section{Related work}
\label{sec:related_work}
\smallskip\noindent\textbf{Data Minimization.}
Existing research on minimization in the context of data protection regulations can be broadly divided into \textit{breadth-based} and \textit{depth-based} techniques. Breadth-based strategies aim to minimize data by limiting the number of features~\citep{rastegarpanah2021auditing} or introducing feature generalization~\citep{goldsteen2021data,staab2024principle}. On the other hand, depth-based approaches focus on reducing the number of unique data points by using methods like data pruning~\citep{paul2021deep,sorscher2022beyond,shanmugam2022learning}. 
While there are some discussions on individualized minimization for recommender systems~\citep{biega2020operationalizing,chen2023studying}, they are limited in their ability to generalize to other settings in ML.

On a separate note, most studies in data minimization aim to simply reduce the raw size of the datasets~\citep{rastegarpanah2021auditing,shanmugam2022learning,biega2020operationalizing,chen2023studying} and don't give any attention to privacy concerns~\citep{leemann2022prefer}. Although some works do go beyond dataset size and discuss other aspects of information leakage~\citep{goldsteen2021data}, they still lack connections with real-world privacy risks. The work closest to ours is a concurrent work by \citet{staab2024principle}, which also introduces real-world privacy attacks to quantify privacy leakage after minimization. However, unlike our approach, \citet{staab2024principle} concentrates on breadth-based methods, thus missing the individualized nature of minimization.

Some studies have also formalized data minimization during inference, emphasizing the personalized nature of the process and delving into its privacy implications~\citep{tran2023personalized,james2023participatory}. However, data minimization during inference is distinctly different from data minimization during training, which is the primary focus of our paper.


\smallskip\noindent\textbf{Privacy Auditing with Inference Attacks.}
Privacy auditing often involves the application of inference attacks to assess real-world privacy leakage. These attacks serve as powerful tools to expose potential vulnerabilities and by simulating realistic scenarios~\citep{shokri2017membership,henriksen2016re,yeom2018privacy}, auditors can evaluate the effectiveness of privacy protection mechanisms and identify areas where sensitive information may be revealed. 

\section{Theoretical analysis of baseline techniques}
\label{sec:app_baseline}
We provide the theoretical properties of the model learned on minimized data for feature selection and individualized random subsampling in this section. 
\subsection{Feature selection}
As introduced in the main text, the feature selection framework sorts the importance of features based on their importance in learning task and then remove the least important features. We denote $S\in [p]$ to be the subset of the weakest features that will be not used. The following Theorem \ref{thm:featurSelection} provides the Bayes Mean Squared Error (MSE) when using all features $[p]$ and using a subset of features $[p]\setminus S$.

\begin{theorem}
\label{thm:featurSelection}
Suppose all input features and labels are jointly Gaussian, i.e., $[\bm{x},y] \sim \mathcal{N}(\mu, \Sigma)$, where $\Sigma = \begin{bmatrix} \Sigma_{\bm{x},\bm{x}}, \Sigma_{\bm{x},y} \\ \Sigma_{y,\bm{x}}, \Sigma_{y,y} \end{bmatrix}$. Furthermore, we assume that all input features are mutually independent, i.e., the covariance matrix $\Sigma_{\bm{x},\bm{x}} = diag([\sigma^2_i]^p_{i=1})$ is a diagonal matrix and $\sigma^2_i = Var[\bm{x}^i]$ the variance of $i^{th}$ feature. Then the Bayes MSE when using all input  features $[p]$ and using a subset of input features in  $[p] \setminus S$ in turn are: $Var[y] - \sum^{p}_{i=1}\frac{(Cov(y, \bm{x}^i))^2}{\sigma^2_i} ]$
and  $Var[y] - \sum_{i \in [p] \setminus S}\frac{Cov(y, \bm{x}^i))^2}{\sigma^2_i}$
\end{theorem}

Theorem \ref{thm:featurSelection} suggests data minimization procedure based on feature selection introduces an additional MSE of $ \sum^{p}_{i=1}\frac{(Cov(y, \bm{x}^i))^2}{\sigma^2_i}  - \sum_{i \in [p] \setminus S}\frac{Cov(y, \bm{x}^i))^2}{\sigma^2)_i} = \sum_{i \in S} \frac{(Cov(y, \bm{x}^i))^2}{\sigma^2_i}   $. As long as all features in the removed feature set $ S$ have small correlation with the label, i.e. $Cov(y, \bm{x}^i)\approx 0$, such additional MSE is neligible. 

\begin{proof}
Our proof relies on the properties of multivariate Gaussian variables. In particular, when feature and label are jointly Gaussian, $[\bm{x},y] \sim \mathcal{N}(\mu, \Sigma)$ then  for any subset of features $A \in [p]$ we can derive the following conditional density of label $y$ given partial input features  $\bm{x}_A$\cite{bishop2006pattern}:

$$ P(y|x_A) = \mathcal{N}\big(\mu_{y} + \Sigma_{\bm{x}_A,y} \Sigma^{-1}_{y,y} \Sigma_{y,\bm{x}}, \Sigma_{y,y} - \Sigma_{\bm{x}_A,y}\Sigma^{-1}_{\bm{x},\bm{x}} \Sigma_{y,\bm{x}_A}\big) .$$

In the above equation, $\Sigma_{y,y}$ is the variance of the label $y$, i.e., $\Sigma_{y,y} = \text{Var}(y)$, while  $\Sigma_{\bm{x}_A,y}$ or $\Sigma_{y, \bm{x}_A}$ is the covariance between $y$ and a subset of features $\bm{x}_A$. The Bayes MSE for Gaussian distribution is just the conditional variance \cite{bayes_error}. Hence the Bayes MSE when using all features, i,e $A=[p]$ is: $\Sigma_{y,y} - \Sigma_{y,\bm{x}}\Sigma^{-1}_{\bm{x},\bm{x}} \Sigma_{\bm{x},y}$. By the assumption that the input features are mutually independent $\Sigma_{\bm{x},\bm{x}} = \text{diag}([\sigma^2_i]^p_{i=1})$, the Bayes MSE using all features can be further reduced as: $\text{Var}[y] - \Sigma_{y,\bm{x}}  \text{diag}([\frac{1}{\sigma^2_i}]^p_{i=1})\Sigma_{\bm{x},y} = \text{Var}[y] -\sum_{i \in [p]} \frac{ (Cov(y,\bm{x}_i))^2}{\sigma^2_i}$.

Similarly, we can derive the Bayes MSE when using a subset of features $d\setminus S$ as: $\text{Var}[y] -\sum_{i \in [d]\setminus S} \frac{ (Cov(y,\bm{x}_i))^2}{\sigma^2_i}$.
\end{proof}

\subsection{Personalized random subsampling}

As introduced in the main text, the personalized random subsampling method works by randomly setting the entries of the minimization $\bm{B}_{ij}=\bot$ with a probability $k_p$, which controls the sparsity of the minimized dataset (Here, $k_p = \frac{k}{np}$). This can also be rewritten as randomly selecting a subset $S \in \{ (i,j) | i \in [n], j \in [p]]\}$ of a given size $|S| = np - k$, and remove the entries in $S$. To understand the theoretical behaviour of this method, we first consider the optimal model parameter $\btheta^*(\bX) = \argmin_{\btheta} J(\btheta; \bX, \bY)$ as a function of the data $\bX$. Using this notation, the model learnt on minimized data can be represented as $\hat{\theta} = \theta^*(\bX')$ while the model parameter learned on original data $\optimal{\btheta} = \theta^*(\bX) $. We then have the following Lemma \ref{lem:1} that derives the gradient of the optimal model parameter $\theta^*(\bX)$ w.r.t data $\bX$


\begin{lemma}[Sensitivity of model parameter w.r.t input $\bX$]
\label{lem:1}

Assume the loss function $J(\btheta; \bX, \bY)$ is differentiable w.r.t $\btheta$ and $\bX$, suppose $\btheta^*(\bX) = \argmin_{\btheta} J(\btheta; \bX, \bY) $ then the following holds:
\begin{equation}
\frac{\partial \btheta^*(\bX)}{\partial \bX} = -H^{-1} G,
\end{equation}
where $H=\frac{\partial^2 J(\btheta^*(\bX);\bX, \bY)}{\partial \btheta^2}$ is the Hessian matrix of the loss w.r.t model parameter, while $G= \frac{\partial^2 J(\btheta^*(\bX);\bX, \bY)}{\partial \btheta, \partial \bX}$ is the second derivative of the loss w.r.t model parameter and input data. 
\end{lemma}

\begin{proof}
Since the model parameter $\btheta^*(\bX) = \argmin_{\btheta} J(\btheta; \bX, \bY)$ hence the gradient of the loss $J(.)$ w.r.t  model parameter vanishes at $\btheta^*(\bX)$. In other words 
$$
\frac{\partial J(\btheta^*(\bX); \bX, \bY) }{\partial \btheta} =\bm{0}^T.
$$ 

Take the derivative w.r.t $\bX$ on both sides of the above equation, it  follows:
\begin{equation}\frac{ \partial}{\partial \bX}\frac{\partial J(\btheta^*(\bX); \bX, \bY) }{\partial \btheta}=\bm{0}^T.
\label{eq:zero}
\end{equation}

The L.H.S can be rewritten as 
\begin{align*}\frac{\partial}{\partial \btheta}\frac{\partial J(\btheta^*(\bX); \bX, \bY) }{\partial \bX}& =\frac{\partial}{\partial \btheta}\bigg[\frac{\partial J(\theta^*(X); \bX,\bY)}{\btheta}\frac{\partial \btheta^*(\bX)}{\partial \bX}  
+ \frac{\partial J(\btheta^*(X); \bX,\bY) }{\partial \bX}  \frac{\partial \bX}{\partial \bX}\bigg]\\
&=\frac{\partial^2 J(\btheta^*(\bX); \bX, \bY)}{\partial \btheta^2}\frac{\partial \btheta^*(\bX)}{\partial \bX} + \frac{\partial^2 J(\btheta^*(\bX); \bX, \bY)}{\partial \btheta \partial \bX},
\end{align*}
where the first equation is due to the chain rule. If we put $H =\frac{\partial^2 J(\btheta^*(\bX); \bX, \bY)}{\partial \btheta^2} $ (the Hessian matrix) and $G =\frac{\partial^2 J(\btheta^*(\bX); \bX, \bY)}{\partial \btheta \partial \bX} $, then the L.H.S of Equation \ref{eq:zero} can be rewritten as:

$$ H \frac{\partial \btheta^*(\bX)}{\partial \bX} + G =\bm{0}^T,$$

which implies $\frac{\partial \btheta^*(\bX)}{\partial \bX} = -H^{-1}G$.

\end{proof}

Lemma \ref{lem:1} tells us how much the optimal model parameter changes when the data input changes. Based on this Lemma \ref{lem:1} we can prove the following Lemma \ref{lem:2} regarding the target utility.

\begin{lemma}[Sensitivity of target utility w.r.t input $\bX$]
\label{lem:2}
Given the same settings and conditions as in Lemma \ref{lem:1}, then the following holds:
\begin{equation}
\frac{\partial J(\btheta^*(\bX), \bX, \bY)}{\partial \bX} = -L H^{-1} G,
\end{equation}
where $H$ and $G$ were provided in Lemma \ref{lem:1}, and $L = \frac{\partial J(\btheta^*(\bX), \bX, \bY) }{\partial \btheta}$ is the gradient of loss w.r.t model parameter.

\end{lemma}

\begin{proof}

The proof is based directly on the chain rule:
$$ \frac{\partial J(\btheta^*(\bX), \bX, \bY)}{\partial \bX}  =\frac{\partial J(\btheta^*(\bX), \bX, \bY)}{\partial \btheta^*(bX)} \frac{ \btheta^*(bX)}{\partial \bX} = -LH^{-1}G,$$

where the last equation is by Lemma \ref{lem:1}.
\end{proof}

Based on Lemma \ref{lem:2} we have the following Theorem \ref{thm:p_sub_sampling} that derives the target utility  $J(\hat{\btheta}; \bX, \bY) $ bound based on the original utiltiy $J(\optimal{\btheta}; \bX, \bY) $ as follows:

\begin{theorem}
\label{thm:p_sub_sampling}
Consider the personalized random subsampling framework, in which all features in a random set $S \in \{ (i,j) | i \in [n], j \in [p]\}$ is removed to form the minimized dataset $\bX'$, then the following holds:
\begin{equation}
J(\hat{\btheta};\bX, \bY) \leq J(\optimal{\btheta};\bX, \bY) + \sqrt{2 |S|} \| \bX\|_{\infty} \| LH^{-1}G\|_2 + \mathcal{O}\big( |S| \| X\|^2_{\infty}\big).
\end{equation}
\end{theorem}

\begin{proof}
The proof relies on the first-order Taylor approximation. First, we consider both optimal model parameters $\hat{\btheta}, \optimal{\btheta}$ as a function of the input data, i.e., $\hat{\btheta} = \btheta^*(\bX')$ and $\optimal{\btheta} = \btheta^*(\bX)$. Based on the first-order Taylor approximation around $\bX$ it follows that:

\begin{align}
\begin{split}
J(\btheta^*(\bX'); \bX ,\bY) & \approx  J(\btheta^*(\bX);\bX, \bY) +  (\bX'-\bX)^T\frac{\partial J(\btheta^*(\cX), \bX ,\bY) }{ \partial \bX}  +\mathcal{O}\big( \| \bX' -\bX\|^2_2\big)
\label{eq:infty_norm}
\end{split}
\end{align}

By Lemma \ref{lem:2}, $\frac{\partial J(\btheta^*(\cX), \bX ,\bY) }{ \partial \bX}  = -LH^{-1}G$. Furthermore 
\begin{align}
\begin{split}
\| \bX' -\bX\|^2_2& = \sum^n_{i=1}\sum^p_{j=1} (\bX'_{i,j} -\bX_{i,j})^2 = \sum_{(i,j) \in S} (\bX'_{i,j} -\bX_{i,j})^2 \\
& \leq \sum_{(i,j) \in S} 2 \| \bX\|^2_{\infty} = 2 |S| \|\bX\|^2_{\infty} ,
\label{eq:50}
\end{split}
\end{align}
where the second equation is due to the fact we only remove features in $S$ while the other entries are kept the same. The  inequality is due to the fact that $|\bX'_{i,j} -\bX_{i,j} | \leq 2 \max_{i,j} |\bX_{i,j}| = 2 \| \bX \|_{\infty}  $, since the the imputed data is in the range $ \bX'_{i,j}   \in [\min_{i,j} \bX_{i,j}, \max_{i,j}  \bX_{i,j}]$.

By Equation \ref{eq:infty_norm} and Cauchy-Schwarz inequality for vectors it follows that:
\begin{align} 
\begin{split}
(\bX'-\bX)^T\frac{\partial J(\btheta^*(\cX), \bX ,\bY) }{ \partial \bX} & \leq \| \bX'-\bX \|_2 \| -LH^{-1}G\|_2\\
& \leq \sqrt{2 |S|} \|\bX\|_{\infty}\| LH^{-1}G\|_2.
\label{eq:6}
\end{split}
\end{align}

Applying the results from Equation \ref{eq:50} and Equation \ref{eq:6} to Equation \ref{eq:infty_norm} we verify the correctness of the statement.

\end{proof}

\section{Data minimization algorithms}
\label{sec:app_algorithms}

We provide additional details on various algorithms used in the paper.

\subsection{Baseline techniques}

\smallskip\noindent\textbf{Feature Selection ~\cite{blum1997selection}.} 
Feature selection is a breadth-based minimization strategy that retains only the most important features in the data. The algorithm works by first sorting the features of the dataset in order of their importance,  using a pre-established criterion, to identify a subset $S$ of the least important features. The minimized dataset $\bX'$ is formed from $\bX$ by removing all features in $S$. 
In our paper, the importance criterion is the absolute correlation between each feature and the output label, and the algorithm sets:
\begin{equation}
    \bm{B}_{ij} = \bot \quad \forall i \in [n], j \in  S
\end{equation}
where $|S|$ is a parameter controlling the minimization sparsity.

\smallskip\noindent\textbf{Random Subsampling.} 
Random subsampling is a depth-based minimization strategy that randomly chooses a subset of data points from the original dataset. In the context of data minimization, it sets the minimization matrix $\bm{B}$ as:
\begin{align}
    \bm{B}_{ij} = \bot \quad \forall j \in [p], \textrm{ with probability }\; k_p,
\end{align}
where $0 \leq k_p \leq 1$ is a probability value chosen so that $nk_p$ rows of the dataset $D$ are minimized, in expectation.

\smallskip\noindent\textbf{Individualized Random Subsampling.} 
Individualized random subsampling is an extension of random subsampling, but it removes individual entries (feature, sample) rather than complete rows, i.e., it performs individualized minimization.
The minimization matrix $\bm{B}$ is defined as:
\begin{align}
    \bm{B}_{ij} = \bot \quad \textrm{ with probability }\; k_p,
\end{align}
where $0 \leq k_p \leq 1$ is chosen so that $npk_p$ elements of the dataset $D$ are minimized, in expectation.

\subsection{Data minimization algorithms}
\label{sec:data_min_alg}

\smallskip\noindent\textbf{Approximating the Lower Level Program~\cite{6079692}.}
Solving the bi-level optimization discussed is challenging due to the nested structure of the problem which requires solving a non-convex lower-level optimization within a non-convex upper-level optimization.
When the lower-level problem has a unique solution that can be explicitly expressed in a closed form, then the overall bilevel program can be rewritten as a single-level program which is much simpler to solve. The underlying idea behind the proposed framework lies in approximating the target utility via the original utility by the first Taylor approximation (see again Equation \ref{eq:infty_norm}). Assuming that the second order component associated with $\| \bX' -\bX\|^2_2$ is negligible, the difference in model's utility is 
$$
J(\hat{\theta}; \bX, \bY) - J(\optimal{\theta};\bX,\bY) \approx -(\bX'-\bX)^T L H^{-1} G.
$$
where $L, H, G$ are the gradients w.r.t.~the model's parameters, the Hessian w.r.t.~the model's parameters, and the second-order derivative w.r.t.~the model's parameters and the dataset, respectively, of the original utility on the complete dataset $J(\optimal{\theta}; \bX, \bY)$. This simplifies the original optimization as:
\begin{equation}
\label{eq:integer_program}
    \minimize_{\bm{B} \in \{\bot, 1\}^{n \times p}} \;\;
    \| \bm{B} \|_1 \quad \quad
        \textsl{s.t.}:
        \;\; (\bX-\bX')^T L H^{-1} G \leq \alpha.
\end{equation}
Suppose, we perform a simple zero imputation, i.e., setting $ \bot= 0$,  then $\bX' = \bX \odot \bm{B}$. The above problem now is a binary integer linear programming and the dual problem can be rewritten as below:

\begin{subequations}
\label{eq:integer_program_approx}
    \begin{align}
    \minimize_{\bm{B} \in \{\bot, 1\}^{n \times p}} \;\; &  (\bX \odot \bm{B}-\bX)^T L H^{-1} G  \\
        \textsl{s.t.}:&\;\;  \| \bm{B} \|_1 \leq k .
    \end{align}
\end{subequations}
Recall again that it is often convenient to view the optimization expressed in \eqref{eq:dm_goal} as a program that minimizes the loss $J(\hat{\theta}; \mathbf{X}, \mathbf{Y})$ under sparsity constraint over the minimization matrix $\bm{B}$.
It turns out that for this binary linear programming, we can obtain a closed-form solution by setting $\bm{B}_{ij} =1$ for $k$ largest entries of  $(\bX_{ij} -\bX'_{ij}) \odot (L H^{-1} G)_{ij}$, and $\bm{B}_{ij} = \bot$ otherwise. Note that this proposed method is applicable when the imputed data is given (e.g., zero imputation) before running the method. Complicated data imputation like mean/mode/median imputation by the non-missing entries of the same column in the minimized data will result in a non-linear objective in the dual problem. This poses a difficulty since there is no efficient method to solve the binary integer non-linear programming in general.

\smallskip\noindent\textbf{Modeling the Target Utility \cite{wang2006review}.}
Instead of relying on the assumptions of low sparsity to approximate the target utility as above, we can take a more general approach by directly modelling and learning the mapping between the target utility and the minimized dataset. In other words, we want to learn a parametrized function $m_\omega(\bm{B}) \approx J(\hat{\theta}; \bX, \bY)$, where $\omega$ is a vector of parameters, to estimate the target utility without solving the lower-level optimization for $\hat{\theta}$.

To learn a tractable mapping $m_\omega(\bm{B})$, we restrict ourselves to linear functions of $\bm{B}$ and assume that each index of the dataset has an independent influence on the target utility. Therefore, if we quantify the influence of each index on the target utility as $\cI_{ij}$, the mapping $m_\omega(\bm{B})$ becomes:
\begin{equation}
    J(\hat{\theta}; \bX, \bY) \approx m_\omega(\bm{B}) = \frac{1}{n p} \sum^n_{i=1} \sum^p_{j=1} \cI_{ij}.
\end{equation} 
Here $\cI_{ij}$ can only exist in one of two binary states, i.e., $\cI_{ij} = \mathbbm{1}_{[\bm{B}_{ij}=1]} \cdot \cI^1_{ij} + \mathbbm{1}_{[\bm{B}_{ij}=\bot]} \cdot \cI^{\bot}_{ij}$. To learn the values of $\cI^1_{ij}$ and $\cI^{\bot}_{ij}$, we generate a large number of minimization matrix-target utility pairs by solving the lower-level optimization for each pair. We can then learn these parameters for each index ${ij}$ by averaging the target utility when $\bm{B}_{ij}=1$ and $\bot$, respectively. The final optimization thus becomes:
\begin{equation}
\label{eq:metamodel}
    \minimize_{\bm{B} \in \{\bot, 1\}^{n \times p}} \;\; 
    \| \bm{B} \|_1 \quad \quad
        \textsl{s.t.}:
        \;\; \frac{1}{n p} \sum^n_{i=1} \sum^p_{j=1} \cI_{ij} - J(\optimal{\theta}; \bX, \bY) \leq \alpha.
\end{equation}
When considering a sparsity constraint $\|\bm{B}\|_1 \leq k$ on the minimization matrix, the solution to this formulation is retaining the $k$ entries with the highest value of the term $\cI^{\bot}_{ij} - \cI^1_{ij}$. Although the assumption 
of independent influence may not hold in all cases, 
our results (illustrated in the next section) show that this approach can substantially improve the accuracy of existing baselines. 

We need to generate minimization matrix-target utility pairs.
We start by generating a random minimization matrix $\bm{B}$, with the target sparsity $k$. This is the same as performing personalized random subsampling. We then solve the lower-level program for this $\bm{B}$ and obtain the final target utility of the trained model. We repeat these steps multiple times to create a dataset to learn the parameters of the mapping $m_\omega(\bm{B})$.

To learn the mapping $m_\omega(\bm{B})$, we simply need to learn the parameters $\cI^1_{ij}, \cI^{\bot}_{ij}$. Given the assumption of the independent influence of each index on the minimization matrix, we can learn these parameters by calculating the average loss when $B_{ij}=1$ and $B_{ij}=\bot$:
\begin{align}
    \cI^1_{ij} = \frac{1}{|P^1|} \sum^{|P^1|}_{q=1} J(\hat{\theta}_{P^1_q}; \cX, \cY) \quad \textrm{where} \quad P^1 := \{\bm{B} \, | \, \bm{B}_{ij}=1\} \\
    \cI^{\bot}_{ij} = \frac{1}{|P^{\bot}|} \sum^{|P^{\bot}|}_{q=1} J(\hat{\theta}_{P^{\bot}_q}; \cX, \cY) \quad \textrm{where} \quad P^{\bot} := \{\bm{B} \, | \, \bm{B}_{ij}=\bot\}\\
    \textrm{where} \quad \hat{\theta}_{P} = 
                \argmin_\theta \frac{1}{n} \sum_{i=1}^n 
                \ell\left(f_\theta( \bm{x}_i \odot P_i), y_i\right).
\end{align}

\smallskip\noindent\textbf{Evolutionary Algorithms \cite{sinha2014finding}.}
So far, we have discussed methods that approximate the original optimization problem under various assumptions, enabling us to solve it more easily. An orthogonal class of algorithms, often applied to solve bi-level programs, are evolutionary algorithms. They trade the advantage of carrying no assumption with slow convergence, i.e., high computational demand, and higher risks of overfitting. 

In our implementation of the evolutionary algorithm, we begin with a population of randomly generated minimization matrices $\bm{B}$ and then evolve them across iterations to reach our objective. We mutate and breed the current population at each stage of the process to create a larger pool of choices and carry out the lower-level optimization for every member of this pool, only retaining the best performers for the next generation. The entire process is repeated until convergence, or for a fixed number of iterations. 

\paragraph{Mutation:} In our paper, we mutate a minimization matrix $\bm{B}$ by flipping the value of exactly 10 randomly chosen indices from $1 \rightarrow \bot$, and exactly 10 randomly chosen indices from $\bot \rightarrow 1$.

\paragraph{Breeding:} When breeding between two parent minimization matrices $\bm{B}^1$ and $\bm{B}^2$, we keep the same value in the child $\bm{B}^c$ at indices where both parents agree to be the same, while we randomly choose values for indices where they don't agree to maintain target sparsity. In simpler terms,
\begin{align}
    \bm{B}^c_{ij} \Leftarrow \bm{B}^1_{ij} \textrm{ if }\; \bm{B}^1_{ij} = \bm{B}^2_{ij},\\
    \bm{B}^c_{ij} \Leftarrow \bot \textrm{ with probability }\; k', \textrm{ otherwise }
\end{align}
where $0 \leq k' \leq 1$ is a value chosen so that the sparsity $k$ is maintained, in expectation. 

\subsection{Summary}

We provide a summary of various strengths and weaknesses of all algorithms used in our paper in Table \ref{tab:multiplicity}.

\begin{table*}[hbtp!]
    \centering
    \footnotesize
    \begin{tabular}{p{0.15\linewidth}p{0.25\linewidth}p{0.25\linewidth}p{0.25\linewidth}}
        \toprule
         & Approximating Lower-Level Program & Modeling Target Utility & Evolutionary Algorithms \\
        \midrule
        Assumptions & All errors between the original and minimized data beyond the first order are considered insignificant and ignored. This might not hold when sparsity is high. & The influence of the presence or absence of every value in the data is modelled independently. This can break for datasets with highly correlated features. & No assumptions are made about the structure of the problem setting. \\
        \\
        Hyperparameter \newline Sensitivity & No hyperparameters. & Large number of iterations required to create better data when modelling the lower optimization. & Large number of iterations as well as a large active population size will facilitate better solutions. \\
        \\
        Consistency & No randomness in the process, but it inherits the randomness of the lower-level learning model. & Highly inconsistent across changing randomness. But consistent across sparsity, i.e. data minimized at lower sparsity will also be minimized at higher sparsity. & Highly inconsistent across both randomness and sparsity. Data minimized may not remain minimized under changing randomness or increasing sparsity. \\
        \\
        Convergence Behavior & Provides an exact solution under the given assumptions. & No convergence guarantees. & Convergence guarantees under a sufficiently large number of iterations. \\
        \\
        Runtime Considerations & Closed form solution, but requires second-order derivatives. Fast for simpler settings, but does not scale well with either dataset size or model complexity. & Requires training the learning model multiple times. Relatively slower for simpler settings, but scales better with increasing complexity. & Requires training the learning model a significantly larger number of times. Furthermore, needs to be repeated for every unique output sparsity required.\\
        \\
        Selection Criteria & Choose in simple settings with a small amount of minimization required for fast and accurate results. & Choose in more complex settings and when algorithm runtime is as important as the accuracy of the method. & Choose in a setting where the accuracy of the method is of the utmost importance, even at the cost of compute. \\
        \bottomrule
    \end{tabular}
    \caption{A summary of strengths and weaknesses of various algorithms.}
    \label{tab:multiplicity}
\end{table*}

\section{Membership inference attacks and data minimization}
\label{app:mia}

In this section, we focus on threat models outside the ``wall'', specifically addressing inference attacks on the trained model without direct access to the minimized dataset.

\subsection{Membership inference risk and inference attack}

\noindent\textbf{Membership Inference Risk (MIR).}
Membership inference attacks~\cite{shokri2017membership} aim to discern if an individual's data was in the dataset before minimization, focusing on an attacker accessing the model $\hat{\theta}$ trained on this minimized dataset. In these attacks, the adversary calculates a likelihood score $L(\bm{x}_q)$ for each query $\bm{x}_q$, representing its probability of being in the original dataset $\bX$ using the model $\hat{\theta}$. The score is given by $L(\bm{x}_q) = \Pr\left[\bm{x}_q \in \bX \, \vert \, \hat{\theta}\right]$. Using these scores for both original dataset $\bX$ and non-members $\bX_{nm}$, binary membership predictions can be made at any threshold $t$, denoted as $\mathbbm{1}_{L(\bm{x}_q) \geq t}$. The overall Membership Inference Risk (MIR) is assessed as the area under the curve (AUC) of true positive rates ($\overrightarrow{\mathrm{tpr}}$) and false positive rates ($\overrightarrow{\mathrm{fpr}}$) across various thresholds:
\begin{align}
    \textrm{MIR} = \textrm{AUC}(\overrightarrow{\mathrm{tpr}}, \overrightarrow{\mathrm{fpr}})
\end{align}

\noindent\textbf{Attack Details.}
The attacker aims to determine the presence or absence of an individual in the original training dataset. We use the SOTA membership inference attack RMIA~\citep{zarifzadehLowCostHighPowerMembership2023}, by training $8$ reference models on random $50\%$ subsets of the public data split. We operate under the practical assumption that the adversary is unaware of the data minimization applied before training the target model and evaluate the membership inference on the original dataset.

\subsection{Privacy leakage through membership inference}

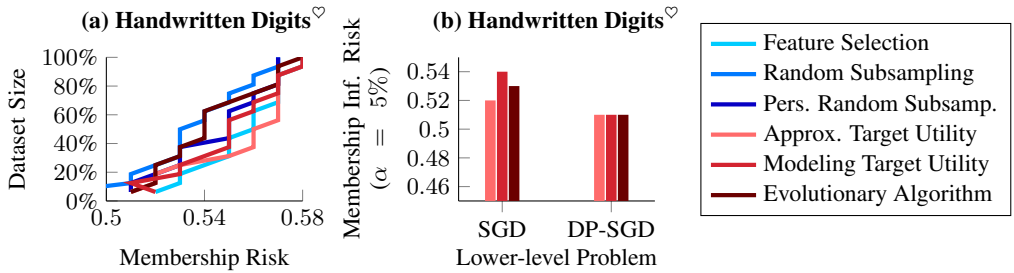
\begin{figure}
    \centering
    \footnotesize
    \begin{tikzpicture}
\begin{axis} [width=0.3\linewidth,height=0.25\linewidth,yticklabel= {\pgfmathprintnumber\tick\%},ymin=0,ymax=100,xticklabel= {\pgfmathprintnumber\tick},xmin=0.5,xmax=0.58,xlabel={Membership Risk},ylabel={Dataset Size},axis y line*=left,axis x line*=bottom,ylabel near ticks,ylabel style={align=center, text width=4cm},xlabel style={align=center},xtick={0.5,0.54,0.58},legend pos=south east, legend style={font=\tiny}, legend cell align={left}, title={\textbf{(a) Handwritten Digits$^\heartsuit$}}]

\addplot [featsel,no marks,ultra thick] table[x=MIA,y=Size] {data/digits_nn/feature_selection.dat};
\addplot [randsub,no marks,ultra thick] table[x=MIA,y=Size] {data/digits_nn/random_subsampling.dat};
\addplot [perrandsub,no marks,ultra thick] table[x=MIA,y=Size] {data/digits_nn/personalized_random_subsampling.dat};
\addplot [taylor,no marks,ultra thick] table[x=MIA,y=Size] {data/digits_nn/taylor_approx_heuristic.dat};
\addplot [genetic,no marks,ultra thick] table[x=MIA,y=Size] {data/digits_nn/genetic_heuristic.dat};
\addplot [metamodel,no marks,ultra thick] table[x=MIA,y=Size] {data/digits_nn/metamodeling_heuristic.dat};

\end{axis}
\end{tikzpicture}%
\begin{tikzpicture}
\begin{axis}[width=0.3\linewidth,height=0.25\linewidth,,ylabel={Membership Inf. Risk \\($\alpha=5\%$)},ymin=0.45,ymax=0.55,xlabel={Lower-level Problem}, enlarge x limits=0.4,xtick=data,ybar=0.5pt,bar width=4pt,axis y line*=left,axis x line*=bottom,symbolic x coords={{SGD},{DP-SGD}},ylabel near ticks,x tick label style  = {text width=2cm,align=center},y label style  = {text width=4cm,align=center},title={\textbf{(b) Handwritten Digits$^\heartsuit$}}
]

\addplot[style={fill=taylor},draw=none
]
coordinates {({SGD},0.52)
             ({DP-SGD},0.51)
	     };
\addplot[style={fill=metamodel},draw=none
]
coordinates {({SGD},0.54)
             ({DP-SGD},0.51)
	     };
\addplot[style={fill=genetic},draw=none
]
coordinates {({SGD},0.53)
             ({DP-SGD},0.51)
	     };

\end{axis}
\end{tikzpicture}%
\raisebox{3em}{
\begin{tikzpicture}
\begin{axis}[scale=0.01,
legend cell align={left},
hide axis,
xmin=0, xmax=1,
ymin=0, ymax=1,
legend columns=1,
legend style={font=\footnotesize,/tikz/every even column/.append style={column sep=0.1cm}}
]
\addlegendimage{ultra thick, featsel}
\addlegendentry{Feature Selection};
\addlegendimage{ultra thick, randsub}
\addlegendentry{Random Subsampling};
\addlegendimage{ultra thick, perrandsub}
\addlegendentry{Pers. Random Subsamp.};
\addlegendimage{ultra thick, taylor}
\addlegendentry{Approx. Target Utility};
\addlegendimage{ultra thick, metamodel}
\addlegendentry{Modeling Target Utility};
\addlegendimage{ultra thick, genetic}
\addlegendentry{Evolutionary Algorithm};

\end{axis}
\end{tikzpicture}
}
    \caption{Membership inference risks under changing sparsity on the handwritten digits dataset. The minimization algorithms can reduce inference risks and are pushed to even better trade-offs by introducing DP-SGD in the lower-level objective.}
    \label{fig:mia}
\end{figure}

Finally, we assess the membership inference risk under various minimization algorithms in Figure \ref{fig:mia}(a). As previously mentioned, information leakage through a trained model is not an expected benefit of data minimization, which mainly aims to address data breach scenarios. Nevertheless, we still observe that certain minimization algorithms are effective at reducing membership inference risks with decreasing dataset size, i.e., models trained on minimized datasets leak less information. Yet, these improvements are not perfectly aligned, carrying forward the same trends we saw in the two data breach scenarios above.

\subsection{DP-SGD to improve privacy-utility trade-off}

We analyze modifications in data minimization to counter membership inference. 
An effective mitigation can be obtained by introducing a differentially private learning algorithm, DP-SGD~\citep{abadi2016deep}, into the lower-level program of the bi-level optimization in \eqref{eq:dm_goal}. We test its compatibility with data minimization by re-evaluating membership inference in this new setting. Note, we do not introduce DP-SGD into utility calculation, i.e., once the data is minimized, the rest of the pipeline outside the ``wall'' remains unchanged.

\smallskip\noindent\textbf{Evaluating DP-SGD Modifications.} 
By incorporating DP-SGD into the minimization optimization without altering other components, we want to assess the compatibility between these two methods. 
The results, collected in Figure \ref{fig:mia}(b), clearly demonstrate a reduction in membership inference risks at the same accuracy threshold, with prominent improvements for methods that were more susceptible to information leakage, such as modelling target utility algorithms.
DP-SGD is indeed compatible with minimization, reinforcing the benefits of considering privacy during minimization.

\section{Additional experiments}
\label{sec:app_defense}
\subsection{Results on additional datasets}
\label{sec:app_add_data}

We also provide results for utility and privacy across changing sparsity on additional datasets. This includes the wine quality dataset~\cite{cortez2009modeling}, a dataset containing various attributes of 6,463 unique wines and a binary label to classify them as red or white wine, and the ACSIncome and ACSEmployment tasks of the folktables dataset~\cite{ding2021retiring}, with census information about individuals and a label marking whether their income is above $\$50,000$ for ACSIncome or whether they are employed or not for ACSEmployment. Similar to 20 newsgroup dataset, we only choose a random subset of 5000 data points from both ACSIncome and ACSEmployment datasets. We also provide detailed results on the bank dataset, the handwritten digits dataset, and the 20 newsgroup dataset, in this section.

\begin{figure*}[bt!]
    \centering
    \footnotesize\input{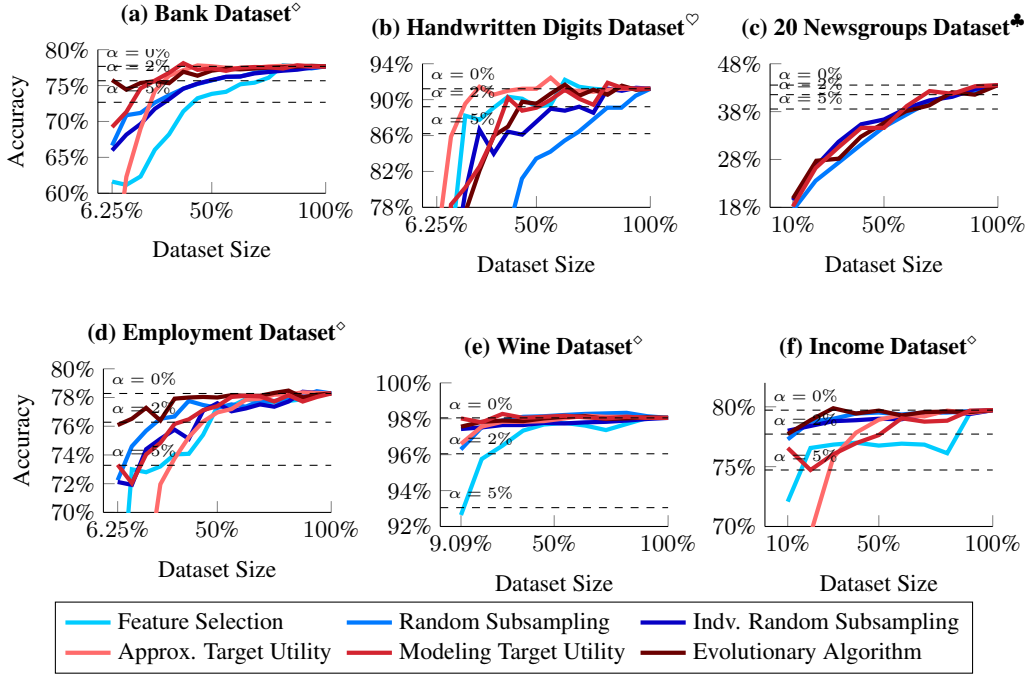}
    \begin{tikzpicture}
\begin{axis}[scale=0.01,
legend cell align={left},
hide axis,
xmin=0, xmax=1,
ymin=0, ymax=1,
legend columns=3,
legend style={font=\footnotesize,/tikz/every even column/.append style={column sep=0.1cm}}
]
\addlegendimage{ultra thick, featsel}
\addlegendentry{Feature Selection};
\addlegendimage{ultra thick, randsub}
\addlegendentry{Random Subsampling};
\addlegendimage{ultra thick, perrandsub}
\addlegendentry{Indv. Random Subsampling};
\addlegendimage{ultra thick, taylor}
\addlegendentry{Approx. Target Utility};
\addlegendimage{ultra thick, metamodel}
\addlegendentry{Modeling Target Utility};
\addlegendimage{ultra thick, genetic}
\addlegendentry{Evolutionary Algorithm};

\end{axis}
\end{tikzpicture}
    \caption{Utility of the minimized data across various sparsities on all datasets.}
    \label{fig:app_dataset_size}
\end{figure*}

\begin{figure*}[bt!]
    \centering
    \footnotesize\begin{tikzpicture}
\begin{axis} [width=0.33\linewidth,height=0.25\linewidth,yticklabel= {\pgfmathprintnumber\tick\%},ylabel={Dataset Size},ymin=0,ymax=100,xticklabel= {\pgfmathprintnumber\tick},xmin=0,xmax=1,xlabel={Re-identification Risk},axis y line*=left,axis x line*=bottom,ylabel near ticks,ylabel style={align=center, text width=4cm},legend pos=south east, legend style={font=\tiny}, legend cell align={left}, title={\textbf{(a) Employment Dataset$^\diamond$}}]

\addplot [featsel,no marks,ultra thick] table[x=Reid,y=Size] {data/employment/feature_selection.dat};
\addplot [randsub,no marks,ultra thick] table[x=Reid,y=Size] {data/employment/random_subsampling.dat};
\addplot [perrandsub,no marks,ultra thick] table[x=Reid,y=Size] {data/employment/personalized_random_subsampling.dat};
\addplot [taylor,no marks,ultra thick] table[x=Reid,y=Size] {data/employment/taylor_approx_heuristic.dat};
\addplot [genetic,no marks,ultra thick] table[x=Reid,y=Size] {data/employment/genetic_heuristic.dat};
\addplot [metamodel,no marks,ultra thick] table[x=Reid,y=Size] {data/employment/metamodeling_heuristic.dat};


\end{axis}
\end{tikzpicture}%
\begin{tikzpicture}
\begin{axis} [width=0.33\linewidth,height=0.25\linewidth,yticklabel= {\pgfmathprintnumber\tick\%},ymin=0,ymax=100,xticklabel= {\pgfmathprintnumber\tick},xmin=0,xmax=1,xlabel={Re-identification Risk},axis y line*=left,axis x line*=bottom,ylabel near ticks,ylabel style={align=center, text width=4cm},legend pos=south east, legend style={font=\tiny}, legend cell align={left}, title={\textbf{(b) Wine Dataset$^\diamond$}}]

\addplot [featsel,no marks,ultra thick] table[x=Reid,y=Size] {data/wine/feature_selection.dat};
\addplot [randsub,no marks,ultra thick] table[x=Reid,y=Size] {data/wine/random_subsampling.dat};
\addplot [perrandsub,no marks,ultra thick] table[x=Reid,y=Size] {data/wine/personalized_random_subsampling.dat};
\addplot [taylor,no marks,ultra thick] table[x=Reid,y=Size] {data/wine/taylor_approx_heuristic.dat};
\addplot [genetic,no marks,ultra thick] table[x=Reid,y=Size] {data/wine/genetic_heuristic.dat};
\addplot [metamodel,no marks,ultra thick] table[x=Reid,y=Size] {data/wine/metamodeling_heuristic.dat};


\end{axis}
\end{tikzpicture}%
\begin{tikzpicture}
\begin{axis} [width=0.33\linewidth,height=0.25\linewidth,yticklabel= {\pgfmathprintnumber\tick\%},ymin=0,ymax=100,xticklabel= {\pgfmathprintnumber\tick},xmin=0,xmax=1,xlabel={Re-identification Risk},axis y line*=left,axis x line*=bottom,ylabel near ticks,ylabel style={align=center, text width=4cm, font=\small},legend pos=south east, legend style={font=\tiny}, legend cell align={left}, title={\textbf{(c) Income Dataset$^\diamond$}}]

\addplot [featsel,no marks,ultra thick] table[x=Reid,y=Size] {data/income/feature_selection.dat};
\addplot [randsub,no marks,ultra thick] table[x=Reid,y=Size] {data/income/random_subsampling.dat};
\addplot [perrandsub,no marks,ultra thick] table[x=Reid,y=Size] {data/income/personalized_random_subsampling.dat};
\addplot [taylor,no marks,ultra thick] table[x=Reid,y=Size] {data/income/taylor_approx_heuristic.dat};
\addplot [genetic,no marks,ultra thick] table[x=Reid,y=Size] {data/income/genetic_heuristic.dat};
\addplot [metamodel,no marks,ultra thick] table[x=Reid,y=Size] {data/income/metamodeling_heuristic.dat};


\end{axis}
\end{tikzpicture}
    \begin{tikzpicture}
\begin{axis} [width=0.33\linewidth,height=0.25\linewidth,yticklabel= {\pgfmathprintnumber\tick\%},ylabel={Dataset Size},ymin=0,ymax=100,xticklabel= {\pgfmathprintnumber\tick},xmin=0,xmax=1,xlabel={Reconstruction Risk},axis y line*=left,axis x line*=bottom,ylabel near ticks,ylabel style={align=center, text width=4cm},legend pos=south east, legend style={font=\tiny}, legend cell align={left}, title={\textbf{(d) Employment Dataset$^\diamond$}}]

\addplot [featsel,no marks,ultra thick] table[x=Recon,y=Size] {data/employment/feature_selection.dat};
\addplot [randsub,no marks,ultra thick] table[x=Recon,y=Size] {data/employment/random_subsampling.dat};
\addplot [perrandsub,no marks,ultra thick] table[x=Recon,y=Size] {data/employment/personalized_random_subsampling.dat};
\addplot [taylor,no marks,ultra thick] table[x=Recon,y=Size] {data/employment/taylor_approx_heuristic.dat};
\addplot [genetic,no marks,ultra thick] table[x=Recon,y=Size] {data/employment/genetic_heuristic.dat};
\addplot [metamodel,no marks,ultra thick] table[x=Recon,y=Size] {data/employment/metamodeling_heuristic.dat};


\end{axis}
\end{tikzpicture}%
\begin{tikzpicture}
\begin{axis} [width=0.33\linewidth,height=0.25\linewidth,yticklabel= {\pgfmathprintnumber\tick\%},ymin=0,ymax=100,xticklabel= {\pgfmathprintnumber\tick},xmin=0,xmax=1,xlabel={Reconstruction Risk},axis y line*=left,axis x line*=bottom,ylabel near ticks,ylabel style={align=center, text width=4cm},legend pos=south east, legend style={font=\tiny}, legend cell align={left}, title={\textbf{(e) Wine Dataset$^\diamond$}}]

\addplot [featsel,no marks,ultra thick] table[x=Recon,y=Size] {data/wine/feature_selection.dat};
\addplot [randsub,no marks,ultra thick] table[x=Recon,y=Size] {data/wine/random_subsampling.dat};
\addplot [perrandsub,no marks,ultra thick] table[x=Recon,y=Size] {data/wine/personalized_random_subsampling.dat};
\addplot [taylor,no marks,ultra thick] table[x=Recon,y=Size] {data/wine/taylor_approx_heuristic.dat};
\addplot [genetic,no marks,ultra thick] table[x=Recon,y=Size] {data/wine/genetic_heuristic.dat};
\addplot [metamodel,no marks,ultra thick] table[x=Recon,y=Size] {data/wine/metamodeling_heuristic.dat};


\end{axis}
\end{tikzpicture}%
\begin{tikzpicture}
\begin{axis} [width=0.33\linewidth,height=0.25\linewidth,yticklabel= {\pgfmathprintnumber\tick\%},ymin=0,ymax=100,xticklabel= {\pgfmathprintnumber\tick},xmin=0,xmax=1,xlabel={Reconstruction Risk},axis y line*=left,axis x line*=bottom,ylabel near ticks,ylabel style={align=center, text width=4cm, font=\small},legend pos=south east, legend style={font=\tiny}, legend cell align={left}, title={\textbf{(f) Income Dataset$^\diamond$}}]

\addplot [featsel,no marks,ultra thick] table[x=Recon,y=Size] {data/income/feature_selection.dat};
\addplot [randsub,no marks,ultra thick] table[x=Recon,y=Size] {data/income/random_subsampling.dat};
\addplot [perrandsub,no marks,ultra thick] table[x=Recon,y=Size] {data/income/personalized_random_subsampling.dat};
\addplot [taylor,no marks,ultra thick] table[x=Recon,y=Size] {data/income/taylor_approx_heuristic.dat};
\addplot [genetic,no marks,ultra thick] table[x=Recon,y=Size] {data/income/genetic_heuristic.dat};
\addplot [metamodel,no marks,ultra thick] table[x=Recon,y=Size] {data/income/metamodeling_heuristic.dat};


\end{axis}
\end{tikzpicture}
    \begin{tikzpicture}
\begin{axis}[scale=0.01,
legend cell align={left},
hide axis,
xmin=0, xmax=1,
ymin=0, ymax=1,
legend columns=3,
legend style={font=\footnotesize,/tikz/every even column/.append style={column sep=0.1cm}}
]
\addlegendimage{ultra thick, featsel}
\addlegendentry{Feature Selection};
\addlegendimage{ultra thick, randsub}
\addlegendentry{Random Subsampling};
\addlegendimage{ultra thick, perrandsub}
\addlegendentry{Indv. Random Subsampling};
\addlegendimage{ultra thick, taylor}
\addlegendentry{Approx. Target Utility};
\addlegendimage{ultra thick, metamodel}
\addlegendentry{Modeling Target Utility};
\addlegendimage{ultra thick, genetic}
\addlegendentry{Evolutionary Algorithm};

\end{axis}
\end{tikzpicture}
    \caption{Re-identification and reconstruction risks under changing sparsity on additional datasets.}
    \label{fig:app_reid_recon}
\end{figure*}
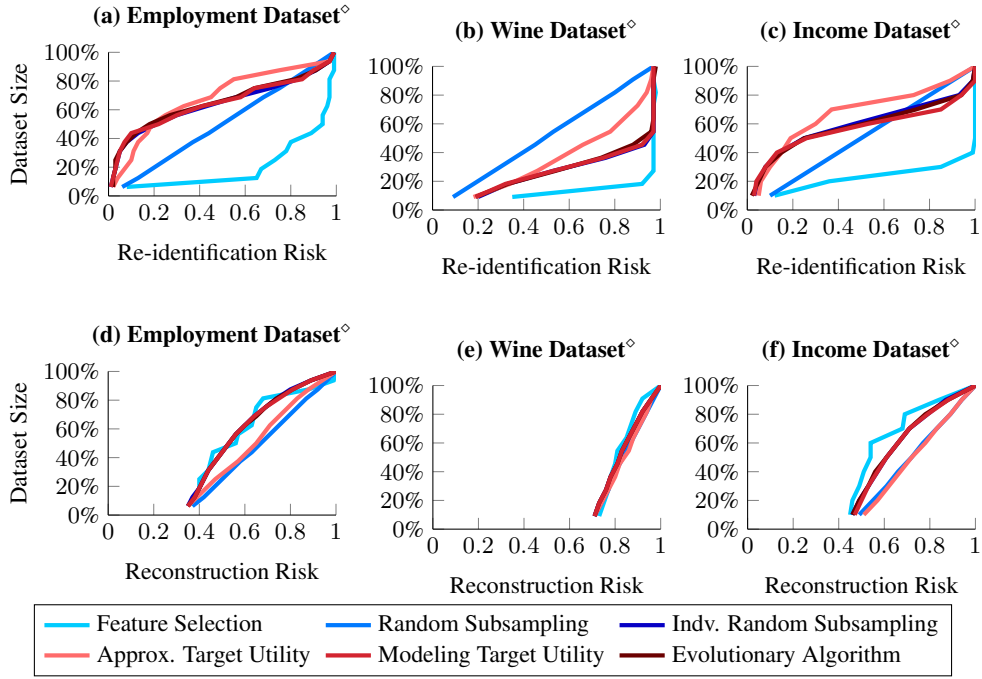

The results for utility and privacy are collected in Figure \ref{fig:app_dataset_size} and Figure \ref{fig:app_reid_recon}, respectively. The trends of utility match the behaviour seen in the main text, i.e., these datasets have redundant information that can be removed without suffering any performance loss. Similarly, for privacy results, we see the trends of main text like feature selection highly misaligned with re-identification, and all methods containing high reconstruction risks even after extreme minimization, replicated in these datasets. Thus, there is a clear misalignment between data minimization and privacy expectations.

\begin{figure*}[hbt!]
    \centering
    \footnotesize
    \begin{tikzpicture}
\begin{axis} [width=0.33\linewidth,height=0.25\linewidth,yticklabel= {\pgfmathprintnumber\tick\%},ylabel={Accuracy},ymin=78,ymax=94,xticklabel= {\pgfmathprintnumber\tick},xmin=0,xmax=1,xlabel={Re-identification Risk},axis y line*=left,axis x line*=bottom,ylabel near ticks,ylabel style={align=center, text width=4cm},xtick={0,0.5,1},ytick={78,82,86,90,94},legend pos=south east, legend style={font=\tiny}, legend cell align={left}]

\addplot [taylordef,no marks,ultra thick] table[x=Reid,y=Acc] {data/digits_nn/taylor_approx_heuristic.dat};
\addplot [taylordef,no marks,ultra thick,dashed] table[x=Reid,y=Acc] {data/digits_nn_reid_1_5/taylor_approx_heuristic.dat};
\addplot [taylordef,no marks,ultra thick,densely dotted] table[x=Reid,y=Acc] {data/digits_nn_reid_3/taylor_approx_heuristic.dat};

\addplot [metamodeldef,no marks,ultra thick] table[x=Reid,y=Acc] {data/digits_nn/metamodeling_heuristic.dat};
\addplot [metamodeldef,no marks,ultra thick,dashed] table[x=Reid,y=Acc] {data/digits_nn_reid_1_5/metamodeling_heuristic.dat};
\addplot [metamodeldef,no marks,ultra thick,densely dotted] table[x=Reid,y=Acc] {data/digits_nn_reid_3/metamodeling_heuristic.dat};

\addplot [geneticdef,no marks,ultra thick] table[x=Reid,y=Acc] {data/digits_nn/genetic_heuristic.dat};
\addplot [geneticdef,no marks,ultra thick,dashed] table[x=Reid,y=Acc] {data/digits_nn_reid_1_5/genetic_heuristic.dat};
\addplot [geneticdef,no marks,ultra thick,densely dotted] table[x=Reid,y=Acc] {data/digits_nn_reid_3/genetic_heuristic.dat};

\draw[black, dashed] (axis cs:0,91.21) -- (axis cs:100,91.21);
\node[anchor=south west] at (axis cs: 0,91.21) {\tiny$\alpha=0\%$};
\draw[black, dashed] (axis cs:0,89.21) -- (axis cs:100,89.21);
\node[anchor=south west] at (axis cs: 0,89.21) {\tiny$\alpha=2\%$};
\draw[black, dashed] (axis cs:0,86.21) -- (axis cs:100,86.21);
\node[anchor=south west] at (axis cs: 0,86.21) {\tiny$\alpha=5\%$};

\end{axis}
\end{tikzpicture}%
\begin{tikzpicture}
\begin{axis} [width=0.33\linewidth,height=0.25\linewidth,yticklabel= {\pgfmathprintnumber\tick\%},ymin=78,ymax=94,xticklabel= {\pgfmathprintnumber\tick},xmin=0.15,xmax=0.4,xlabel={Reconstruction Risk},axis y line*=left,axis x line*=bottom,ylabel near ticks,ylabel style={align=center, text width=4cm},xtick={0.2, 0.3,0.4},ytick={78,82,86,90,94},legend pos=south east, legend style={font=\tiny}, legend cell align={left}]

\addplot [taylordef,no marks,ultra thick] table[x=Recon,y=Acc] {data/digits_nn/taylor_approx_heuristic.dat};
\addplot [taylordef,no marks,ultra thick,dashed] table[x=Recon,y=Acc] {data/digits_nn_recon_1_5/taylor_approx_heuristic.dat};
\addplot [taylordef,no marks,ultra thick,densely dotted] table[x=Recon,y=Acc] {data/digits_nn_recon_3/taylor_approx_heuristic.dat};

\addplot [metamodeldef,no marks,ultra thick] table[x=Recon,y=Acc] {data/digits_nn/metamodeling_heuristic.dat};
\addplot [metamodeldef,no marks,ultra thick,dashed] table[x=Recon,y=Acc] {data/digits_nn_recon_1_5/metamodeling_heuristic.dat};
\addplot [metamodeldef,no marks,ultra thick,densely dotted] table[x=Recon,y=Acc] {data/digits_nn_recon_3/metamodeling_heuristic.dat};

\addplot [geneticdef,no marks,ultra thick] table[x=Recon,y=Acc] {data/digits_nn/genetic_heuristic.dat};
\addplot [geneticdef,no marks,ultra thick,dashed] table[x=Recon,y=Acc] {data/digits_nn_recon_1_5/genetic_heuristic.dat};
\addplot [geneticdef,no marks,ultra thick,densely dotted] table[x=Recon,y=Acc] {data/digits_nn_recon_3/genetic_heuristic.dat};

\draw[black, dashed] (axis cs:0.15,91.21) -- (axis cs:0.4,91.21);
\node[anchor=south west] at (axis cs: 0.15,91.21) {\tiny$\alpha=0\%$};
\draw[black, dashed] (axis cs:0.15,89.21) -- (axis cs:0.4,89.21);
\node[anchor=south west] at (axis cs: 0.15,89.21) {\tiny$\alpha=2\%$};
\draw[black, dashed] (axis cs:0.15,86.21) -- (axis cs:0.4,86.21);
\node[anchor=south west] at (axis cs: 0.15,86.21) {\tiny$\alpha=5\%$};

\end{axis}
\end{tikzpicture}
\hspace{1em}
\raisebox{3em}{
\begin{tikzpicture}
\begin{axis}[scale=0.01,
legend cell align={left},
hide axis,
xmin=0, xmax=1,
ymin=0, ymax=1,
legend columns=1,
legend style={font=\footnotesize,/tikz/every even column/.append style={column sep=0.1cm}}
]
\addlegendimage{ultra thick, taylordef}
\addlegendentry{Approx. Target Utility};
\addlegendimage{ultra thick, metamodeldef}
\addlegendentry{Modeling Target Utility};
\addlegendimage{ultra thick, geneticdef}
\addlegendentry{Evolutionary Algorithm};
\addlegendimage{ultra thick, white}
\addlegendentry{};
\addlegendimage{ultra thick, gray}
\addlegendentry{$\beta=0$};
\addlegendimage{ultra thick, gray, dashed}
\addlegendentry{$\beta=1.5$};
\addlegendimage{ultra thick, gray, densely dotted}
\addlegendentry{$\beta=3$};

\end{axis}
\end{tikzpicture}
}
    \caption{Re-identification and reconstruction risks (zoomed in) on the handwritten digits dataset, using feature-level privacy scores.}
	\label{fig:app_reid_recon_defense_digits}
\end{figure*}
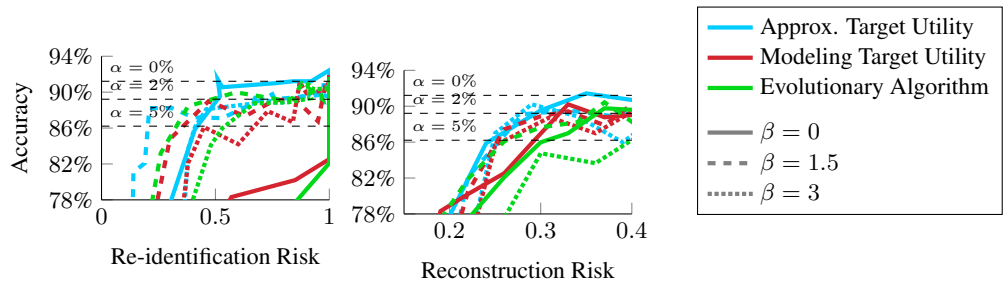

\begin{figure*}[hbt!]
    \centering
    \footnotesize
    \begin{tikzpicture}
\begin{axis} [width=0.33\linewidth,height=0.25\linewidth,yticklabel= {\pgfmathprintnumber\tick\%},ylabel={Accuracy},ymin=60,ymax=80,xticklabel= {\pgfmathprintnumber\tick},xmin=0,xmax=1,xlabel={Re-identification Risk},axis y line*=left,axis x line*=bottom,ylabel near ticks,ylabel style={align=center, text width=4cm},xtick={0,0.5,1},legend pos=south east, legend style={font=\tiny}, legend cell align={left}]

\addplot [taylordef,no marks,ultra thick] table[x=Reid,y=Acc] {data/bank/taylor_approx_heuristic.dat};
\addplot [taylordef,no marks,ultra thick,dashed] table[x=Reid,y=Acc] {data/bank_reid_1_5/taylor_approx_heuristic.dat};
\addplot [taylordef,no marks,ultra thick,densely dotted] table[x=Reid,y=Acc] {data/bank_reid_3/taylor_approx_heuristic.dat};

\addplot [metamodeldef,no marks,ultra thick] table[x=Reid,y=Acc] {data/bank/metamodeling_heuristic.dat};
\addplot [metamodeldef,no marks,ultra thick,dashed] table[x=Reid,y=Acc] {data/bank_reid_1_5/metamodeling_heuristic.dat};
\addplot [metamodeldef,no marks,ultra thick,densely dotted] table[x=Reid,y=Acc] {data/bank_reid_3/metamodeling_heuristic.dat};

\addplot [geneticdef,no marks,ultra thick] table[x=Reid,y=Acc] {data/bank/genetic_heuristic.dat};
\addplot [geneticdef,no marks,ultra thick,dashed] table[x=Reid,y=Acc] {data/bank_reid_1_5/genetic_heuristic.dat};
\addplot [geneticdef,no marks,ultra thick,densely dotted] table[x=Reid,y=Acc] {data/bank_reid_3/genetic_heuristic.dat};

\draw[black, dashed] (axis cs:0,77.67) -- (axis cs:100,77.67);
\node[anchor=south west] at (axis cs: 0,77.67) {\tiny $\alpha=0\%$};
\draw[black, dashed] (axis cs:0,75.67) -- (axis cs:100,75.67);
\node[anchor=south west] at (axis cs: 0,75.67) {\tiny $\alpha=2\%$};
\draw[black, dashed] (axis cs:0,72.67) -- (axis cs:100,72.67);
\node[anchor=south west] at (axis cs: 0,72.67) {\tiny $\alpha=5\%$};

\end{axis}
\end{tikzpicture}%
\begin{tikzpicture}
\begin{axis} [width=0.33\linewidth,height=0.25\linewidth,yticklabel= {\pgfmathprintnumber\tick\%},ymin=60,ymax=80,xticklabel= {\pgfmathprintnumber\tick},xmin=0.35,xmax=0.6,xlabel={Reconstruction Risk (Zoomed In)},axis y line*=left,axis x line*=bottom,ylabel near ticks,ylabel style={align=center, text width=4cm},xtick={0.2,0.3,0.4,0.5,0.6},legend pos=south east, legend style={font=\tiny}, legend cell align={left}]

\addplot [taylordef,no marks,ultra thick] table[x=Recon,y=Acc] {data/bank/taylor_approx_heuristic.dat};
\addplot [taylordef,no marks,ultra thick,dashed] table[x=Recon,y=Acc] {data/bank_recon_1_5/taylor_approx_heuristic.dat};
\addplot [taylordef,no marks,ultra thick,densely dotted] table[x=Recon,y=Acc] {data/bank_recon_3/taylor_approx_heuristic.dat};

\addplot [metamodeldef,no marks,ultra thick] table[x=Recon,y=Acc] {data/bank/metamodeling_heuristic.dat};
\addplot [metamodeldef,no marks,ultra thick,dashed] table[x=Recon,y=Acc] {data/bank_recon_1_5/metamodeling_heuristic.dat};
\addplot [metamodeldef,no marks,ultra thick,densely dotted] table[x=Recon,y=Acc] {data/bank_recon_3/metamodeling_heuristic.dat};

\addplot [geneticdef,no marks,ultra thick] table[x=Recon,y=Acc] {data/bank/genetic_heuristic.dat};
\addplot [geneticdef,no marks,ultra thick,dashed] table[x=Recon,y=Acc] {data/bank_recon_1_5/genetic_heuristic.dat};
\addplot [geneticdef,no marks,ultra thick,densely dotted] table[x=Recon,y=Acc] {data/bank_recon_3/genetic_heuristic.dat};

\draw[black, dashed] (axis cs:0.35,77.67) -- (axis cs:0.6,77.67);
\node[anchor=south west] at (axis cs: 0.35,77.67) {\tiny $\alpha=0\%$};
\draw[black, dashed] (axis cs:0.35,75.67) -- (axis cs:0.6,75.67);
\node[anchor=south west] at (axis cs: 0.35,75.67) {\tiny $\alpha=2\%$};
\draw[black, dashed] (axis cs:0.35,72.67) -- (axis cs:0.6,72.67);
\node[anchor=south west] at (axis cs: 0.35,72.67) {\tiny $\alpha=5\%$};

\end{axis}
\end{tikzpicture}
\hspace{1em}
\raisebox{3em}{
\begin{tikzpicture}
\begin{axis}[scale=0.01,
legend cell align={left},
hide axis,
xmin=0, xmax=1,
ymin=0, ymax=1,
legend columns=1,
legend style={font=\footnotesize,/tikz/every even column/.append style={column sep=0.1cm}}
]
\addlegendimage{ultra thick, taylordef}
\addlegendentry{Approx. Target Utility};
\addlegendimage{ultra thick, metamodeldef}
\addlegendentry{Modeling Target Utility};
\addlegendimage{ultra thick, geneticdef}
\addlegendentry{Evolutionary Algorithm};
\addlegendimage{ultra thick, white}
\addlegendentry{};
\addlegendimage{ultra thick, gray}
\addlegendentry{$\beta=0$};
\addlegendimage{ultra thick, gray, dashed}
\addlegendentry{$\beta=1.5$};
\addlegendimage{ultra thick, gray, densely dotted}
\addlegendentry{$\beta=3$};

\end{axis}
\end{tikzpicture}
}
    \caption{Re-identification and reconstruction risks (zoomed in) on the bank dataset, using feature-level privacy scores.}
	\label{fig:app_reid_recon_defense_bank}
\end{figure*}
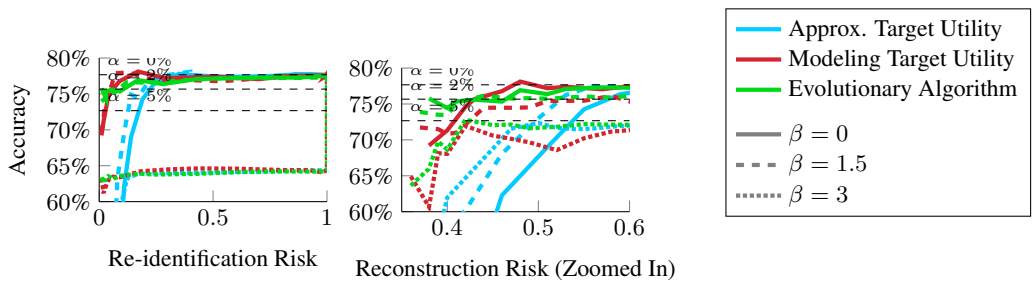

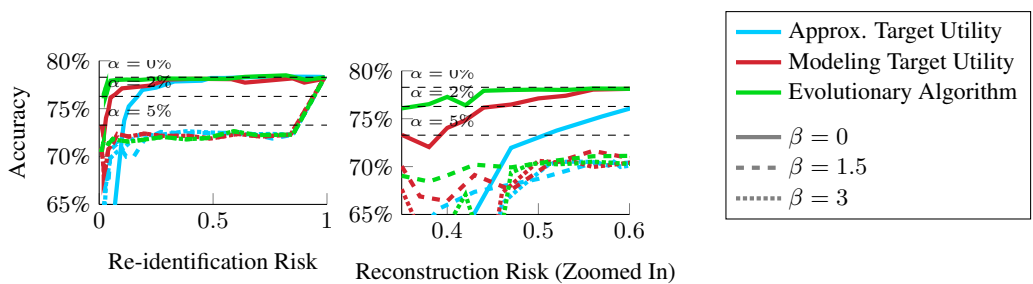
\begin{figure*}[hbt!]
    \centering
    \footnotesize
    \begin{tikzpicture}
\begin{axis} [width=0.33\linewidth,height=0.25\linewidth,yticklabel= {\pgfmathprintnumber\tick\%},ylabel={Accuracy},ymin=65,ymax=80,xticklabel= {\pgfmathprintnumber\tick},xmin=0,xmax=1,xlabel={Re-identification Risk},axis y line*=left,axis x line*=bottom,ylabel near ticks,ylabel style={align=center, text width=4cm},xtick={0,0.5,1},legend pos=south east, legend style={font=\tiny}, legend cell align={left}]

\addplot [taylordef,no marks,ultra thick] table[x=Reid,y=Acc] {data/employment/taylor_approx_heuristic.dat};
\addplot [taylordef,no marks,ultra thick,dashed] table[x=Reid,y=Acc] {data/employment_reid_1_5/taylor_approx_heuristic.dat};
\addplot [taylordef,no marks,ultra thick,densely dotted] table[x=Reid,y=Acc] {data/employment_reid_3/taylor_approx_heuristic.dat};

\addplot [metamodeldef,no marks,ultra thick] table[x=Reid,y=Acc] {data/employment/metamodeling_heuristic.dat};
\addplot [metamodeldef,no marks,ultra thick,dashed] table[x=Reid,y=Acc] {data/employment_reid_1_5/metamodeling_heuristic.dat};
\addplot [metamodeldef,no marks,ultra thick,densely dotted] table[x=Reid,y=Acc] {data/employment_reid_3/metamodeling_heuristic.dat};

\addplot [geneticdef,no marks,ultra thick] table[x=Reid,y=Acc] {data/employment/genetic_heuristic.dat};
\addplot [geneticdef,no marks,ultra thick,dashed] table[x=Reid,y=Acc] {data/employment_reid_1_5/genetic_heuristic.dat};
\addplot [geneticdef,no marks,ultra thick,densely dotted] table[x=Reid,y=Acc] {data/employment_reid_3/genetic_heuristic.dat};

\draw[black, dashed] (axis cs:0,78.28) -- (axis cs:100,78.28);
\node[anchor=south west] at (axis cs: 0,78.28) {\tiny$\alpha=0\%$};
\draw[black, dashed] (axis cs:0,76.28) -- (axis cs:100,76.28);
\node[anchor=south west] at (axis cs: 0,76.28) {\tiny$\alpha=2\%$};
\draw[black, dashed] (axis cs:0,73.28) -- (axis cs:100,73.28);
\node[anchor=south west] at (axis cs: 0,73.28) {\tiny$\alpha=5\%$};

\end{axis}
\end{tikzpicture}%
\begin{tikzpicture}
\begin{axis} [width=0.33\linewidth,height=0.25\linewidth,yticklabel= {\pgfmathprintnumber\tick\%},ymin=65,ymax=80,xticklabel= {\pgfmathprintnumber\tick},xmin=0.35,xmax=0.6,xlabel={Reconstruction Risk (Zoomed In)},axis y line*=left,axis x line*=bottom,ylabel near ticks,ylabel style={align=center, text width=4cm},xtick={0.2,0.3,0.4,0.5,0.6},legend pos=south east, legend style={font=\tiny}, legend cell align={left}]

\addplot [taylordef,no marks,ultra thick] table[x=Recon,y=Acc] {data/employment/taylor_approx_heuristic.dat};
\addplot [taylordef,no marks,ultra thick,dashed] table[x=Recon,y=Acc] {data/employment_recon_1_5/taylor_approx_heuristic.dat};
\addplot [taylordef,no marks,ultra thick,densely dotted] table[x=Recon,y=Acc] {data/employment_recon_3/taylor_approx_heuristic.dat};

\addplot [metamodeldef,no marks,ultra thick] table[x=Recon,y=Acc] {data/employment/metamodeling_heuristic.dat};
\addplot [metamodeldef,no marks,ultra thick,dashed] table[x=Recon,y=Acc] {data/employment_recon_1_5/metamodeling_heuristic.dat};
\addplot [metamodeldef,no marks,ultra thick,densely dotted] table[x=Recon,y=Acc] {data/employment_recon_3/metamodeling_heuristic.dat};

\addplot [geneticdef,no marks,ultra thick] table[x=Recon,y=Acc] {data/employment/genetic_heuristic.dat};
\addplot [geneticdef,no marks,ultra thick,dashed] table[x=Recon,y=Acc] {data/employment_recon_1_5/genetic_heuristic.dat};
\addplot [geneticdef,no marks,ultra thick,densely dotted] table[x=Recon,y=Acc] {data/employment_recon_3/genetic_heuristic.dat};

\draw[black, dashed] (axis cs:0.35,78.28) -- (axis cs:0.6,78.28);
\node[anchor=south west] at (axis cs:0.35,78.28) {\tiny$\alpha=0\%$};
\draw[black, dashed] (axis cs:0.35,76.28) -- (axis cs:0.6,76.28);
\node[anchor=south west] at (axis cs:0.35,76.28) {\tiny$\alpha=2\%$};
\draw[black, dashed] (axis cs:0.35,73.28) -- (axis cs:0.6,73.28);
\node[anchor=south west] at (axis cs:0.35,73.28) {\tiny$\alpha=5\%$};

\end{axis}
\end{tikzpicture}
\hspace{1em}
\raisebox{3em}{
\begin{tikzpicture}
\begin{axis}[scale=0.01,
legend cell align={left},
hide axis,
xmin=0, xmax=1,
ymin=0, ymax=1,
legend columns=1,
legend style={font=\footnotesize,/tikz/every even column/.append style={column sep=0.1cm}}
]
\addlegendimage{ultra thick, taylordef}
\addlegendentry{Approx. Target Utility};
\addlegendimage{ultra thick, metamodeldef}
\addlegendentry{Modeling Target Utility};
\addlegendimage{ultra thick, geneticdef}
\addlegendentry{Evolutionary Algorithm};
\addlegendimage{ultra thick, white}
\addlegendentry{};
\addlegendimage{ultra thick, gray}
\addlegendentry{$\beta=0$};
\addlegendimage{ultra thick, gray, dashed}
\addlegendentry{$\beta=1.5$};
\addlegendimage{ultra thick, gray, densely dotted}
\addlegendentry{$\beta=3$};

\end{axis}
\end{tikzpicture}
}
    \caption{Re-identification and reconstruction risks (zoomed in) on ACSEmployment, using feature-level privacy scores.}
	\label{fig:app_reid_recon_defense_employment}
\end{figure*}

\subsection{Additional results for privacy-based modifications}

We provide additional results for privacy-based modifications to the data minimization algorithm on other datasets, as well as the raw trends on the handwritten dataset. The results for the handwritten dataset are collected in Figure \ref{fig:app_reid_recon_defense_digits}, for the bank dataset are collected in Figure \ref{fig:app_reid_recon_defense_bank}, and for the employment dataset are collected in Figure \ref{fig:app_reid_recon_defense_employment}. The results show similar improvement as in the main text for the bank dataset, however, we don't see similar benefits on the employment dataset. We believe that's because these results are quite sensitive to the choice of hyperparameter $\beta$, and a more thorough search for $\beta$ can show improvements for other datasets as well. Despite this discrepancy, the aim of these modifications was not to propose a novel method of incorporating privacy in minimization, but instead to highlight that minimization and privacy are compatible, and thus one can perform data minimization in line with the regulations while making sure they respect individual privacy in the dataset.


\end{document}